\documentclass[10pt]{article}

\usepackage{yinwang_ads}

\usepackage{bm}
\usepackage{commath}
\usepackage{relsize}
\usepackage{bbm}
\usepackage{nicefrac}
\usepackage{mysymbol}

\usepackage{algorithm}
\usepackage{algpseudocode}
\usepackage{listings}
\usepackage[shortlabels]{enumitem}
\usepackage{pifont}
\usepackage{tikz}
\usetikzlibrary{positioning}
\usepackage{tabularx}
\usepackage{xspace}

\crefname{algorithm}{algorithm}{algorithms}
\Crefname{algorithm}{Algorithm}{Algorithms}

\makeatletter
\renewcommand\author[2][]{\addtolist[#1]{#2}{\yw@authorlist}{\yw@authorformat}{,\hskip 0.5em\relax}}
\makeatother

\hypersetup{
  pdftitle={BeyondEgo: Learning to Act Beyond Egocentric Perception via Humanoid Ego-Exo Foresight}
}

\theoremstyle{definition}

\theoremstyle{remark}

\allowdisplaybreaks

\newcommand{\Method}{BeyondEgo\xspace}
\newcommand{\PsiZero}{\ensuremath{\Psi_0}\xspace}
\newcommand{\sg}{\operatorname{sg}}
\newcommand{\LN}{\operatorname{LN}}
\newcommand{\Dino}{E_{\text{DINOv3}}}
\newcommand{\Lact}{\mathcal{L}_{\mathrm{act}}}
\newcommand{\Lfut}{\mathcal{L}_{\mathrm{fut}}}
\newcommand{\ego}{\mathrm{ego}}
\newcommand{\exo}{\mathrm{exo}}
\newcommand{\Sone}{Feature fusion\xspace}
\newcommand{\Stwo}{VLM fusion\xspace}
\newcolumntype{Y}{>{\raggedright\arraybackslash}X}
\newcolumntype{C}{>{\centering\arraybackslash}X}
\input{results}

\shorttitle{BeyondEgo}

\title{BeyondEgo: Learning to Act Beyond Egocentric\\Perception via Humanoid Ego-Exo Foresight}

\author[1,*]{\protect\mbox{Zhaoshan Yang}}
\author[2,*]{\protect\mbox{Chenglei Li}}
\author[2,*]{\protect\mbox{Haoyu Lei}}
\author[3]{\protect\mbox{Yutao Zhang}}
\author[2]{\protect\mbox{Yuyu Yang}}
\author[2]{\protect\mbox{Zhuo Chen}}
\author[2]{\protect\mbox{Ling Ding}}
\author[2]{\protect\mbox{Zhongxing Ren}}
\author[2]{\protect\mbox{Yi Gu}}
\author[2]{\protect\mbox{Heyang Zhang}}
\author[2]{\protect\mbox{Peng Su}}
\author[4]{\protect\mbox{Jin Pan}}

\affiliation[1]{\protect\mbox{Tongji University}}
\affiliation[2]{\protect\mbox{Yingwang Intelligent Technology}}
\affiliation[3]{\protect\mbox{Universit\protect\'e de Montr\protect\'eal}}
\affiliation[4]{\protect\mbox{The Chinese University of Hong Kong}}

\contribution[*]{Equal contribution}

\abstract{%
Humanoid loco-manipulation requires coordinating whole-body motion to perform manipulation tasks in complex environments. Yet egocentric vision-language-action (VLA) policies can miss critical body--scene relationships, and action imitation alone provides no explicit supervision for future scene evolution. We propose BeyondEgo to extend VLA perception in space and time. Spatially, we encode a synchronized exocentric view with frozen DINOv3 and concatenate the resulting tokens with egocentric VLM tokens to condition the action expert. Temporally, we use future observations from both views as privileged supervision for a residual adapter on task-conditioned VLM features. A foresight expert regresses DINOv3 representations of both views 0.6\,s ahead. At deployment, the foresight expert is discarded and the 4.2M-parameter adapter is retained.
We also introduce HELM, to our knowledge, the first closed-loop humanoid loco-manipulation benchmark with synchronized egocentric and exocentric views across five whole-body tasks.
On HELM, BeyondEgo improves success from 48.0\% to 92.0\% in distribution and from 23.5\% to 65.0\% on 200 held-out scenes with layout, asset, material, and lighting shifts. Ablations show that dual-view DINOv3 foresight yields the best transfer results.%
}

\begin{document}
\maketitle

\begin{figure}[!htb]
\centering
\includegraphics[width=0.94\linewidth]{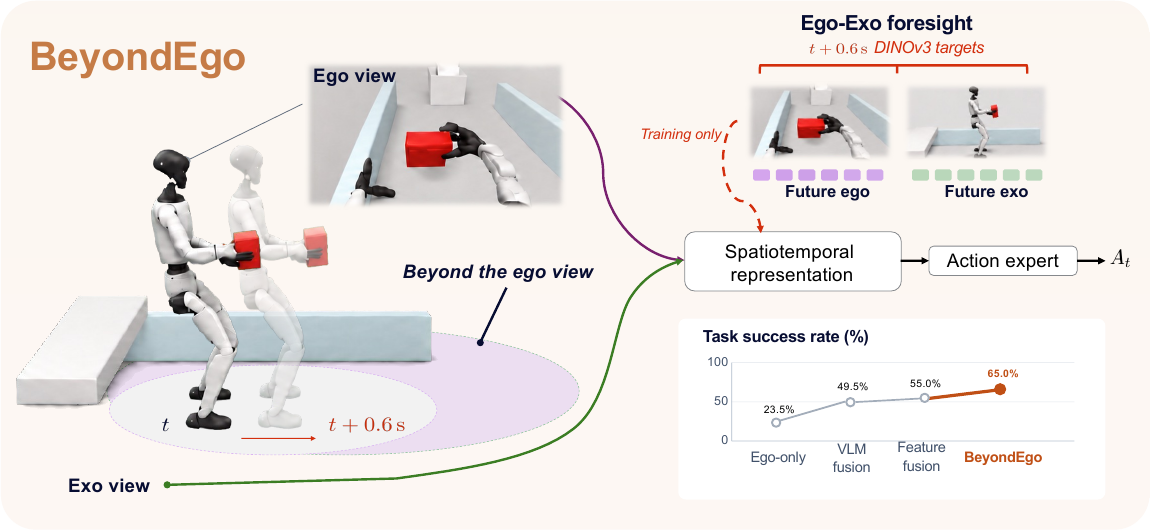}
\caption{\textbf{Learning to act beyond egocentric perception in space and time.}
Egocentric and exocentric observations provide complementary spatial information, while DINOv3 features from both views 0.6\,s ahead provide temporal supervision during training.
\Method integrates these signals into a shared spatiotemporal representation for action generation.
It achieves the highest task success rate among the evaluated methods.}
\label{fig:routes}
\end{figure}


\section{Introduction}
\label{sec:intro}

Humanoid robots offer the potential to extend manipulation beyond fixed workspaces, enabling tasks that require coordinating object interaction with whole-body locomotion. For example, carrying a box through a narrow gap requires maintaining a stable grasp while aligning the torso and limbs with the opening and avoiding surrounding obstacles. Such behaviors depend on understanding the spatial relationship between the robot's body and the environment, as well as anticipating how this relationship will evolve during motion. Recent humanoid vision-language-action (VLA) models make these behaviors increasingly learnable from demonstrations and language instructions, typically by combining a vision-language backbone (System-2), a fast action generator (System-1), and a learned whole-body controller (System-0)~\cite{bjorck2025gr00t,jiang2025wholebodyvla,wei2026psi0}.

Despite this progress, humanoid VLA policies typically rely on head-mounted egocentric cameras, which provide limited cues for whole-body spatial awareness and future-oriented decision-making. The limitation is both spatial and temporal. \emph{Spatially}, the egocentric view moves with the robot and can be occluded by its arms or carried objects, making it difficult to infer critical body--scene relationships, such as the distance between the shoulders and nearby obstacles, the alignment of the body with a narrow opening, or the robot's position relative to a junction. \emph{Temporally}, policies are typically conditioned on current observations without explicit supervision for how these spatial relationships evolve as the robot moves. However, successful loco-manipulation requires anticipating the consequences of ongoing motion, rather than relying solely on what is visible in the current frame. These observations motivate extending the policy's perception beyond the egocentric view in both space and time. No existing benchmark fits this setting: ego-exo datasets are human and offline, while robot multi-view benchmarks target tabletop manipulation.

We propose \Method, which combines exocentric spatial context with dual-view foresight. For the \emph{spatial extension}, we introduce a synchronized view from an additional exocentric camera. We encode this view with frozen DINOv3~\cite{simeoni2025dinov3} and concatenate the resulting tokens with egocentric VLM tokens as input to the action expert (\Cref{fig:routes}). This provides additional visual context about the scene layout and the robot's body relative to its surroundings.
For the \emph{temporal extension}, we use future observations from both views as privileged training supervision~\cite{vapnik2009lupi}. A residual adapter refines VLM latent features before they are passed to the action expert. A foresight expert reads the adapted features together with exocentric features and predicts frozen DINOv3 representations of both views 0.6\,s ahead (\Cref{fig:routes}). The foresight loss supervises the observation interface, while the action expert is trained with flow matching. At deployment, we discard the foresight expert and retain the 4.2M-parameter adapter.

We therefore build HELM, a benchmark of five simulated humanoid loco-manipulation tasks, with 50 paired in-distribution trials and 200 held-out scenes spanning layout, asset, material, lighting, and combined shifts. We evaluate our designs in closed-loop simulation over this benchmark. \Method achieves \BEm{id}{full}{avg}\% in-distribution and \BEm{gen}{full}{avg}\% under distribution shift, improving over both the egocentric \PsiZero baseline (\BEm{id}{psi}{avg}\% and \BEm{gen}{psi}{avg}\%) and the no-foresight \Sone baseline (\BEm{id}{s1}{avg}\% and \BEm{gen}{s1}{avg}\%). Target-space ablations show that predicting DINOv3 features of both views provides the strongest transfer, whereas predicting VLM features does not improve over the no-foresight baseline.

In summary, our contributions are:
\begin{itemize}[leftmargin=1.4em, itemsep=2.5pt, topsep=3pt]

\item A spatial extension to humanoid VLAs through exocentric observations, supported by a comparison of two fusion strategies. We show that action-expert-level fusion provides stronger generalization than VLM-level fusion under the tested distribution shifts.

\item A dual-view foresight objective that uses future observations as privileged supervision to embed future dynamics in VLM latent features. The foresight expert is removed at deployment, while the learned feature adaptation supports action generation with the standard flow-matching objective.

\item \textbf{HELM}, to our knowledge, the first closed-loop humanoid loco-manipulation benchmark with synchronized ego-exo views, featuring five whole-body tasks that require cues beyond the egocentric view, controlled distribution shifts, and per-stage metrics. Our analyses show that the feature space used for fusion and foresight strongly affects closed-loop generalization.

\end{itemize}

\section{Related Work}
\label{sec:related}

\paragraph{Humanoid VLAs and multi-view perception}
Vision-language-action (VLA) models map visual observations and language to robot actions~\cite{brohan2023rt2,kim2024openvla,black2024pi0}, often using flow-matching action experts to generate continuous action chunks~\cite{lipman2023flow,black2024pi0}. Hierarchical architectures combine vision-language backbones (System-2) with fast action generators (System-1)~\cite{bjorck2025gr00t,intelligence2025pi05,wei2026psi0}, while humanoid and legged systems further employ learned whole-body or locomotion controllers~\cite{ding2025humanoidvla,xue2025leverb,cheng2024navila,jiang2025wholebodyvla,wei2026psi0}, supported by teleoperation-based data collection~\cite{he2024omnih2o,fu2024humanplus,li2025amo,ben2025homie}. Recent systems also leverage egocentric human videos and humanoid demonstrations for pre-training~\cite{wei2026psi0,yang2025egovla}, together with real-time action chunking~\cite{black2025rtc,black2025trainingrtc}. Beyond egocentric perception, multi-view manipulation policies combine wrist-mounted and third-person cameras~\cite{zhao2023act,chi2023diffusion,black2024pi0}, with different viewpoints exhibiting distinct generalization trade-offs~\cite{hsu2022seeing}. Synchronized views enable self-supervised learning~\cite{sermanet2018tcn,seo2023mvmwm}, while ego-exo datasets provide paired first- and third-person observations~\cite{sigurdsson2018actor,grauman2024egoexo4d,huang2024egoexolearn}. These capture human activity offline, unlike our robot setting. Geometric representations, such as point clouds, have also been injected directly into VLA action experts~\cite{li2025pointvla,sun2025geovla}. We study exocentric feature fusion in hierarchical humanoid VLAs, exploit future observations from both views as training supervision, and contribute HELM, a benchmark with synchronized ego-exo views and closed-loop evaluation under controlled shifts.

\paragraph{Predictive learning and privileged foresight}
World models predict future observations under actions~\cite{hafner2023dreamerv3,bar2025nwm}, increasingly in latent spaces learned by self-supervised encoders~\cite{zhou2025dinowm,assran2025vjepa2}; self-predictive learning similarly trains encoders to predict future representations~\cite{schwarzer2021spr}. In robot learning, foresight is commonly coupled with action generation, either by conditioning policies on predicted frames, subgoals, or world representations~\cite{du2023unipi,tian2025seer,hu2025vpp,zhao2025cotvla,zhang2025dreamvla}, or by jointly predicting future observations and actions~\cite{wu2024gr1,zhang2025upvla,cen2025worldvla,qin2026wbwam}. Closest to our setting, $\omega$-0~\cite{li2026omega0} predicts future visual latents for humanoid loco-manipulation and accepts either an egocentric or an exocentric view. It uses a single view per task, and its action branch still attends to the future-prediction queries at inference; we fuse both views at every step and remove the foresight expert entirely at deployment. FLARE~\cite{zheng2025flare} aligns policy features with future observation embeddings, while other auxiliary objectives align or reconstruct current inputs using frozen visual features~\cite{yu2025repa,li2025spatialforcing} or image regions~\cite{song2025reconvla}. Learning using privileged information~\cite{vapnik2009lupi} provides another perspective: prior robot learning methods distill privileged simulator states into sensor-based policies~\cite{chen2019lbc,lee2020quadruped,kumar2021rma} or use them in asymmetric critics~\cite{pinto2018asymmetric}. In contrast, \Method treats frozen DINOv3 features of future dual-view frames as privileged regression targets, without requiring simulator states or a state-based teacher policy. Specifically, it supervises the observation interface consumed by the action expert and leaves the action expert's flow-matching objective intact, rather than supervising the policy's internal representations~\cite{zheng2025flare} or sharing a generator between prediction and action~\cite{wu2024gr1,cen2025worldvla}, while discarding the foresight expert at deployment. Our ablations further show that the choice of foresight target space strongly affects closed-loop transfer.

\paragraph{Visual representations and closed-loop evaluation}
Pre-trained visual representations are widely used in robot learning~\cite{nair2022r3m,majumdar2023vc1}, with self-supervised vision transformers~\cite{oquab2024dinov2,simeoni2025dinov3} providing dense features that capture 3D properties such as depth and surface orientation~\cite{elbanani2024probe3d}. Prior VLMs and VLAs fuse such visual features with language-aligned representations inside the backbone~\cite{karamcheti2024prismatic,kim2024openvla}; we instead inject DINOv3 features directly into the action expert and use the same feature space for future supervision. For evaluation, robot policy generalization is increasingly assessed in closed-loop settings under controlled environmental shifts~\cite{pumacay2024colosseum}, with simulation providing useful proxies for real-world performance~\cite{li2024simpler}.

\section{Method: BeyondEgo}
\label{sec:method}

We introduce BeyondEgo, a VLA model for humanoid loco-manipulation. Given a natural language task instruction $\ell$, and synchronized egocentric and exocentric observations together with the robot's proprioceptive state $o_t = (I^{\ego}_t, I^{\exo}_t, \mathbf s_t)$ at time $t$, our model predicts the whole-body action chunk $\mathbf a_{t:t+H}$ of horizon $H$. The action $\mathbf a\in \mathbb R^{36}$ inherits the definition from $\Psi_0$~\cite{wei2026psi0} as $\{\mathbf q_{\mathrm{hand}}, \mathbf q_{\mathrm{arm}}, \mathbf{torso}_{\mathrm{rpy}}, h_{\mathrm{b}}, v_x, v_y, v_{\mathrm{yaw}}, p_{\mathrm{yaw}}\}$, where $\mathbf q_{\mathrm{hand}},\mathbf q_{\mathrm{arm}}\in\mathbb{R}^{14}$, $\mathbf{torso}_{\mathrm{rpy}}\in\mathbb{R}^3$ is the torso roll, pitch and yaw. The remaining five dimensions specify base height, horizontal linear velocities, angular velocity and the target yaw rotation. An RL-based tracking policy executes those commands on the lower-body and torso joints.

\begin{figure}[t]
\centering
\includegraphics[width=\textwidth]{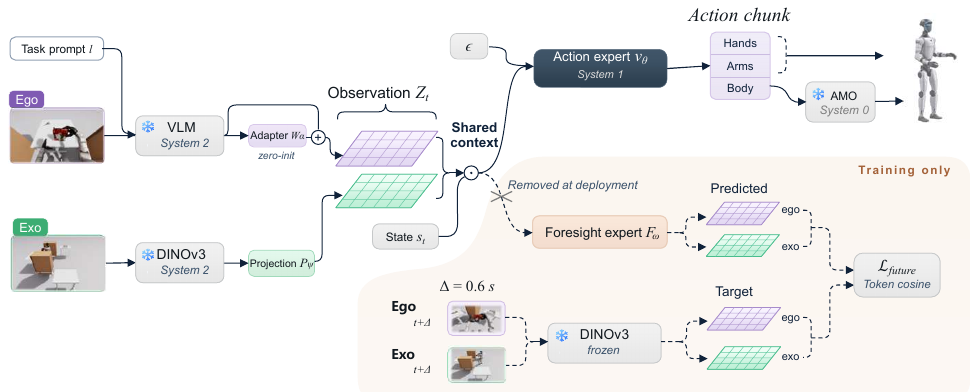}
\caption{\textbf{\Method architecture.}
Frozen VLM and DINOv3 encoders process the ego image with the instruction and the exo image, respectively.
Adapted ego tokens, projected exo tokens, and robot state condition the action expert.
Hand and arm targets are executed directly, while AMO translates body commands into waist and leg joint targets.
During training, the foresight expert predicts dual-view features 0.6\,s ahead, supervised by frozen DINOv3 targets through token-wise cosine loss.
The foresight branch is removed at deployment.}
\label{fig:arch}
\end{figure}

We build on a \emph{triple-system} humanoid VLA model~\cite{wei2026psi0,li2026omega0} (\Cref{sec:prelim}) and extend its perception beyond a single egocentric observation, spatially through an exocentric camera (\Cref{sec:fusion}) and temporally with dual-view foresight (\Cref{sec:foresight}). \Cref{fig:arch} shows the complete model.

\subsection{Model Architecture}
\label{sec:prelim}

Our model organizes humanoid control into three systems~\cite{wei2026psi0,li2026omega0}. System-2 processes observation $o_t$ and provides semantic and geometric understanding through its hidden features. We use Qwen3-VL-2B-Instruct~\cite{bai2025qwen3vl} to encode the egocentric image $I^{\ego}_t$ and language task instruction $\ell$, and DINOv3~\cite{simeoni2025dinov3} to encode the exocentric image $I^{\exo}_t$, providing additional spatial context.
System-1 consists of a foresight expert $F_\omega$ and an action expert. The foresight expert, used only during training, enriches System-2's hidden features with information about future dynamics through a residual block trained with future-frame supervision. The action expert, implemented as a multimodal diffusion transformer (MM-DiT), is trained with flow matching to predict future whole-body action chunks $\mathbf{a}_{t:t+H}$ conditioned on these spatiotemporally augmented features and the proprioceptive states $\mathbf s_t$.
System-0 is an RL-based control policy that converts the predicted actions into robot control commands. Following $\Psi_0$, we use AMO~\cite{li2025amo} to map the lower-body commands $\{\mathbf{torso}_{\mathrm{rpy}}, h_{\mathrm{b}}, v_x, v_y, v_{\mathrm{yaw}}, p_{\mathrm{yaw}}\}$ to leg and waist joint targets.

\subsection{Spatial Extension}
\label{sec:fusion}

To help the model complete challenging tasks that require more spatial context, we introduce a synchronized exocentric observation $I^{\exo}_t$ captured by an additional camera. This view provides \emph{geometric evidence} about the scene layout and the spatial relation between the robot's body and its surroundings, offering additional cues for action prediction. Alongside the VLM, which encodes the egocentric observation $I^{\ego}_t$ and language task instruction $\ell$, we use a frozen DINOv3~\cite{simeoni2025dinov3} encoder to extract spatial context features:
\begin{align}
    Z^{\ego}_t &= E_{\text{VLM}}(I^{\ego}_t, \ell),\\*
    Z^{\exo}_t &= P_\psi(E_{\text{DINOv3}}(I^{\exo}_t)),
\end{align}
where $P_\psi$ is a trainable projection that maps the DINOv3 features to the same embedding dimension as $Z^{\ego}_t$.

An alternative is to feed both images directly into the VLM to produce a unified representation:
\begin{equation}
    Z^{\mathrm{uni}}_t = E_{\text{VLM}}(I^{\ego}_t, I^{\exo}_t, \ell).
\end{equation}
This approach emphasizes joint \emph{semantic understanding} of the two views, but shows suboptimal performance in our experiments. We compare and discuss the two designs in \Cref{sec:where}.

\subsection{Temporal Extension}
\label{sec:foresight}

Standard VLMs are good at visual understanding, but often have limited awareness of how observations evolve over time. To address this limitation, we use future-frame supervision during training to embed information about future dynamics in System-2's encoded features. We supervise predictions in feature space rather than pixel space to focus on high-level dynamics. Specifically, we encode the future egocentric and exocentric frames $I^{\ego}_{t+\Delta t}$ and $I^{\exo}_{t+\Delta t}$ using the same frozen DINOv3 encoder and use the resulting features $\hat{Z}^\ego_{t+\Delta t},\hat{Z}^\exo_{t+\Delta t}$ as prediction targets. We also evaluate predicting future features in Qwen3-VL image feature space and report the comparison in \Cref{sec:what}.

\paragraph{Residual dynamics adapter}
We use a learnable residual adapter~\cite{gao2021clip,houlsby2019param} to incorporate future dynamics into the encoded features. Since only the VLM receives the task instruction and reasons about the intended action, we apply the adapter to its features $Z^{\ego}_t$:
\begin{equation}
    \widetilde{Z}^{\ego}_t = Z^{\ego}_t + W_\alpha(Z^{\ego}_t),
\end{equation}
where $W_\alpha$ is the learned residual branch, zero-initialized to preserve the original features. We then concatenate the adapted features with the unmodified exocentric features $Z^{\exo}_t$ and feed them to the foresight expert $F_\omega$ along with the proprioceptive states $\mathbf s_t$. We jointly train the adapter and $F_\omega$ to predict future features from both views using a cosine-similarity loss:
\begin{equation}
    \label{eq:fut}\Lfut = \mathbb{E}\left[1 - \cos\left([\widehat{Z}^{\ego}_{t+\Delta t};\widehat{Z}^{\exo}_{t+\Delta t}], F_\omega([\widetilde{Z}^{\ego}_t;Z^{\exo}_t], \mathbf s_t)\right)\right].
\end{equation}
At inference, only the learned adapter is kept, and the foresight expert is dropped.

The action expert in System-1 reads the adapted features during both training and inference. The flow-matching training objective is
\begin{equation}
    \label{eq:act}\Lact=\mathbb E\left[\big\|v_\theta([\widetilde{Z}^{\ego}_t;Z^{\exo}_t], \mathbf s_t, \mathbf{a}_t^\sigma, \sigma)-(\boldsymbol\epsilon-\mathbf a_t)\big\|_2^2\right],
\end{equation}
where $\boldsymbol\epsilon$ is Gaussian noise, $\sigma\in[0,1]$ is the flow time, and $\mathbf{a}_t^\sigma = (1-\sigma)\mathbf{a}_t + \sigma\boldsymbol\epsilon$ is the noised action; at test time we integrate $v_\theta$ from $\sigma=1$ to $0$ to produce the action chunk.

\subsection{Implementation and Training Details}
\label{sec:deploy}

\paragraph{Implementation}
We use Qwen3-VL-2B-Instruct~\cite{bai2025qwen3vl} as the VLM backbone in System-2, with feature width $d=2048$. DINOv3 ViT-L/16~\cite{simeoni2025dinov3} extracts features from observation images at $224\times384$ pixels. We retain its class token and apply $2\times2$ average pooling to the patch-token grid, yielding $85$ tokens of dimension $1024$.
The adapter $W_\alpha$ is a single linear projection layer with 4.2M parameters. The foresight expert is a Transformer with hidden dimension $512$ and approximately 19.2M parameters. We set the future prediction horizon $\Delta t$ to 0.6\,s. The System-1 action expert is a 500M-parameter multimodal diffusion transformer with hidden dimension $1536$, trained with flow matching. It predicts chunks of $H=30$ commands for execution at 50\,Hz, covering the same 0.6\,s horizon. We connect modules of different feature widths using learnable MLP projectors.

\paragraph{Exocentric dropout}
To reduce over-reliance on the external view and strengthen the use of egocentric cues, we apply exocentric observation dropout during training. With probability $0.1$, we mask exocentric tokens in all attention blocks and exclude the corresponding exocentric targets from the future prediction loss. This requires the policy to also learn action prediction without access to the external view.

\paragraph{Real-time action chunking}
We adopt the training-time real-time chunking (RTC) procedure in~\cite{black2025trainingrtc}, where they condition each prediction on the previously committed action chunk to maintain consistency across successive chunks. At test time, we use 10 sampling steps and execute the first 15 commands of each chunk before replanning, using RTC~\cite{black2025rtc} with an inference delay of seven control steps.

\paragraph{Training details}
We follow the training recipe of $\Psi_0$~\cite{wei2026psi0} and initialize the VLM and action expert from its released pre-trained and post-trained checkpoints, respectively. The VLM is pre-trained through autoregressive prediction of FAST-tokenized single-step actions, first on EgoDex and then on Humanoid Everyday. The action expert is subsequently post-trained on Humanoid Everyday using flow matching in joint space, with the VLM frozen. For our target tasks, we keep the VLM and DINOv3 frozen and jointly optimize the action expert, residual adapter, foresight expert, and learnable projectors using the action prediction and future-feature supervision losses
\begin{equation}
    \mathcal{L}_{\mathrm{train}} = \mathcal{L}_{\mathrm{act}} + \lambda\mathcal{L}_{\mathrm{fut}},
\end{equation}
where we set $\lambda = 1.0$. We train for 40k steps per task with a global batch size of $128$ on eight B200 GPUs. We use AdamW with $\beta_1=0.95$, $\beta_2=0.999$, and weight decay $10^{-6}$. The learning rate reaches $10^{-4}$ after 1,000 warm-up steps and then follows a cosine decay schedule. Training uses BF16 mixed precision and gradient clipping with a maximum norm of $1.0$. We normalize states and actions using their respective minimum and maximum values and sample flow timesteps uniformly from $[0,1]$. We also apply the training-time RTC procedure described above.

\FloatBarrier
\section{Experiments}
\label{sec:experiments}

We evaluate \Method and the underlying spatial and temporal extensions on five humanoid loco-\hspace{0pt}manipulation tasks, focusing on three questions: How much does the complete method improve task success and generalization? Which spatial fusion strategy and training-time foresight components contribute to these gains? How do the target feature space, supervised views, and exocentric camera placement affect task execution?

\subsection{Experimental Setup}
\label{sec:setup}

\begin{figure}[!t]
\centering
\newcommand{\taskcol}[3]{\begin{minipage}[t]{\dimexpr(\textwidth-8pt)/5\relax}\centering
\parbox[c][2.2\baselineskip][c]{\linewidth}{\centering\fontsize{7.5}{9}\selectfont\sffamily\bfseries #1}\\[2pt]
\includegraphics[width=\linewidth]{#2_exo.jpg}\\[1.5pt]
\includegraphics[width=\linewidth]{#2_ego.jpg}\\[0pt]
{\scriptsize\itshape\strut #3}\end{minipage}}%
\taskcol{T1\\Back Out and Turn}{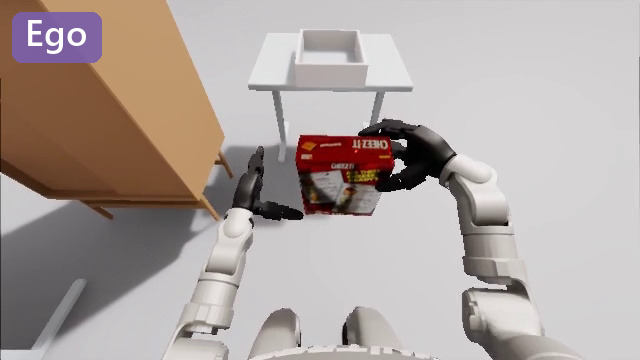}{back out; choose a turn}\hfill
\taskcol{T2\\Turn and Search}{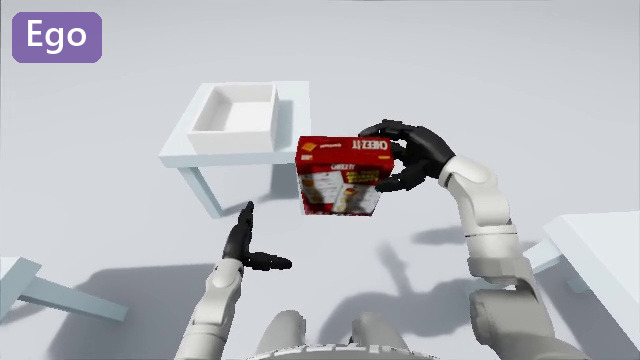}{find the tray behind}\hfill
\taskcol{T3\\Choose the Right Passage}{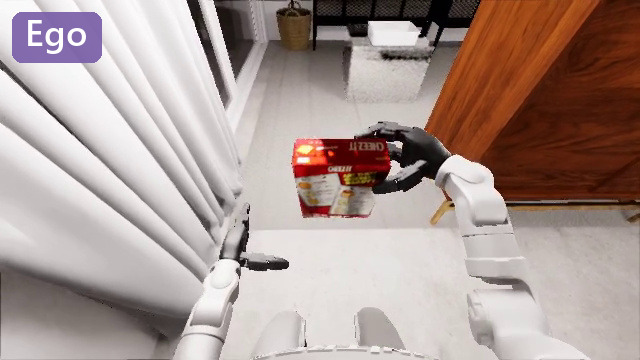}{choose a path around furniture}\hfill
\taskcol{T4\\Sidestep and Turn}{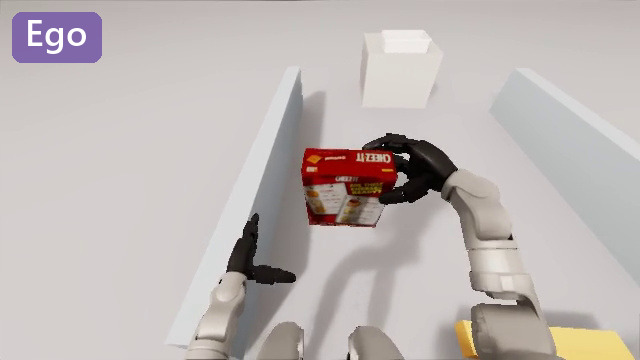}{sidestep before turning}\hfill
\taskcol{T5\\Sideways Passage}{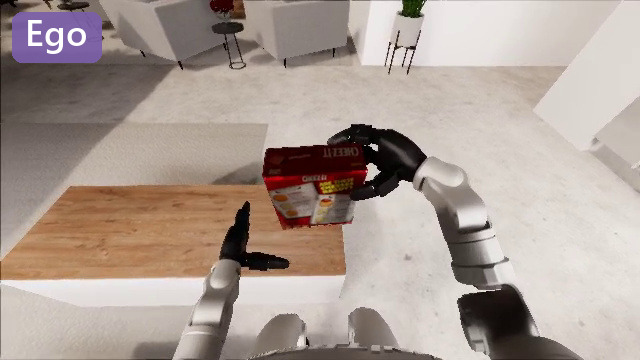}{align and pass sideways}
\caption{\textbf{Five humanoid loco-manipulation tasks.} All tasks require picking up, carrying, and placing an object, but pose different locomotion challenges. Each column shows the robot carrying the box during the demonstration of the corresponding task. The top row shows exocentric images captured by an additional external camera, while the bottom row shows simultaneous egocentric images from a head-mounted camera. The two views provide complementary observations of the carried object, the surrounding environment, and the robot's position relative to obstacles.}
\label{fig:tasks}
\end{figure}

\paragraph{Tasks and data}
We introduce \textbf{HELM} (Humanoid Ego-Exo Loco-Manipulation), a closed-loop benchmark for perception beyond the egocentric view. It comprises five tasks in Isaac Sim on a 29-DoF Unitree G1 with Dex3 hands~(\Cref{fig:tasks}), each built so the decisive cue lies outside or ahead of the head-mounted view. All tasks require picking up a cookie box, moving to the target table, and placing the box into a designated container.
\textbf{T1 (Back Out and Turn)} requires walking backward along a corridor and turning toward the target table at its end; the challenge is to adapt the turning location to the corridor length and select the correct left or right branch. Backing up turns the head camera away from the branch.
\textbf{T2 (Turn and Search)} places the target tray on one of three tables behind the robot, requiring it to turn, locate the tray outside its initial field of view, and approach the correct table. The tray starts behind the robot, out of view.
\textbf{T3 (Choose the Right Passage)} places furniture between the robot and the target table after the robot turns. The furniture occludes the target from the egocentric view, and only one of the two passages around it leads to the table holding the target container. The head camera cannot see which passage is correct.
\textbf{T4 (Sidestep and Turn)} requires a lateral step before turning toward the target table, because turning immediately can cause the robot to trip over low obstacles near its feet. These obstacles sit below the head camera.
\textbf{T5 (Sideways Passage)} requires moving laterally to a narrow passage, aligning with its entrance, and adjusting body orientation to pass through sideways. Body--entrance alignment is hard to see from the head camera.

We develop an automated data pipeline: CuRobo generates arm, navigation, and whole-body control commands, trajectories are rolled out in MuJoCo and recorded at 50\,Hz, and validated trajectories are replayed in Isaac Sim to render synchronized egocentric and exocentric images at $640 \times 360$. In total, 500 valid demonstrations are collected across five tasks.

\paragraph{Evaluation protocols and metrics}
We evaluate in SIMPLE~\cite{simple_sim} (Isaac Sim rendering, MuJoCo physics, AMO motion control). All five tasks have a 35\,s episode limit. A trial succeeds if the robot follows the prescribed route and leaves the box stably inside the target container.
For \emph{in-distribution} evaluation, we sample scenes from the training distribution using a fixed random seed and conduct 10 trials per task, yielding 50 trials in total. For \emph{generalization}, we freeze 40 scenes per task across five shift families (layout, asset, material, lighting, and combined), eight scenes each, 200 in total. All scenes are frozen and hash-pinned, so every method sees identical conditions.
We report per-task and mean success rates, and for T1 the grasp, correct-branch, and placement counts for stage-wise analysis.

\paragraph{Baselines and compared variants}
For all policies, the vision-language model (VLM) and action expert are initialized from the pre-trained weights of \PsiZero. We compare four configurations: the \PsiZero baseline uses only the egocentric view; VLM fusion feeds both views jointly into the VLM; Feature fusion concatenates exocentric features extracted by a frozen DINOv3 encoder with egocentric observation tokens and feeds them directly into the action expert; and BeyondEgo extends Feature fusion with a zero-initialized egocentric adapter and training-only dual-view foresight supervision.
VLM fusion further pre-trains the backbone for approximately 145,000 steps on the existing 500 dual-view demonstrations, then freezes the backbone and fine-tunes the action expert. This reuses the demonstration data but requires additional pre-training computation; Feature fusion does not require this backbone adaptation stage. BeyondEgo sets the foresight loss weight to 1 and uses dual-view visual features $\Delta t=0.6$\,s into the future as supervision targets. The foresight expert is removed after training, while the trained adapter is retained for deployment.

\subsection{Overall Performance and Generalization}
\label{sec:main}

\begin{figure}[t]
\centering
\includegraphics[width=\textwidth]{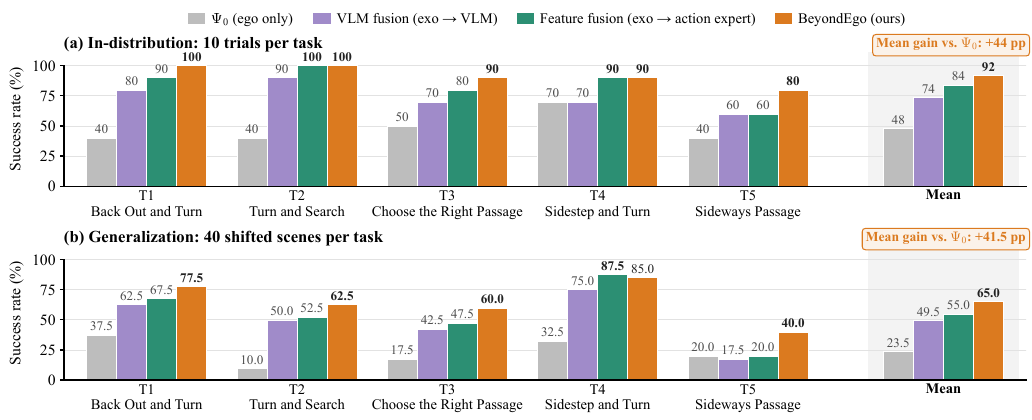}
\caption{\textbf{Task success across five humanoid loco-manipulation tasks.} (a) In-distribution evaluation with 10 trials per task. (b) Generalization evaluation on 40 held-out scenes per task, covering five types of scene variation with eight scenes per type. VLM fusion and Feature fusion introduce the exocentric view through the VLM and the action expert, respectively; BeyondEgo adds training-only dual-view foresight to Feature fusion.}
\label{fig:main}
\end{figure}

\begin{table}[!t]
\centering
\caption{\textbf{Training-seed consistency on T1.} Three independently trained \Method models (training seeds 1234, 5678, and 292285), each evaluated five times on the same 40 shifted generalization scenes (ID: 10 trials); all 15 evaluations fall within 75.0--82.5\%. $\pm$ denotes the SD across the five evaluations of each seed, and across seed means in the last row.}
\label{tab:seeds}
\small
\renewcommand{\arraystretch}{1.16}
\setlength{\tabcolsep}{7pt}
\begin{tabular}{l c ccccc c}
\toprule
\sffamily\bfseries Seed &
\sffamily\bfseries ID (\%) &
\multicolumn{5}{c}{\sffamily\bfseries Generalization (\%)} &
\sffamily\bfseries Mean (\%) \\
\cmidrule(lr){3-7}
 & & \sffamily\bfseries 1 & \sffamily\bfseries 2 & \sffamily\bfseries 3 & \sffamily\bfseries 4 & \sffamily\bfseries 5 & \\
\midrule
Seed 1 & 100.0 & 75.0 & 80.0 & 80.0 & 77.5 & 80.0 & 78.5 $\pm$ 2.2 \\
Seed 2 & 100.0 & 75.0 & 77.5 & 80.0 & 75.0 & 77.5 & 77.0 $\pm$ 2.1 \\
Seed 3 & 100.0 & 75.0 & 82.5 & 80.0 & 80.0 & 77.5 & 79.0 $\pm$ 2.9 \\
\midrule
\ywrowhighlight
\textbf{All seeds} & \textbf{100.0} & \multicolumn{5}{c}{\emph{all 15 evals within 75.0--82.5}} & \textbf{78.2 $\pm$ 1.0} \\
\bottomrule
\end{tabular}
\end{table}

\paragraph{Overall task performance}
BeyondEgo achieves the highest mean success rate among the four configurations in both evaluation suites (\Cref{fig:main}). In distribution, it reaches 92.0\%, improving over the egocentric-only $\Psi_0$ baseline (48.0\%) by 44.0 percentage points. With the exocentric view, VLM fusion and Feature fusion reach 74.0\% and 84.0\%, respectively. BeyondEgo further improves upon the stronger Feature fusion baseline by 8.0 percentage points, demonstrating the performance benefits of combining spatial and temporal extensions.

\paragraph{Generalization to scene variations}
Across 200 generalization scenes, BeyondEgo achieves the highest success rate of 65.0\%, exceeding the 23.5\%, 49.5\%, and 55.0\% achieved by $\Psi_0$, VLM fusion, and Feature fusion, respectively. It succeeds approximately 2.8 times as often as $\Psi_0$ and improves upon Feature fusion by 10.0 percentage points, demonstrating the generalization benefits of the complete method under scene variations.
These gains extend across multiple tasks: BeyondEgo outperforms Feature fusion on four of the five tasks, with the largest improvement on T5 and further gains on T1, T2, and T3. On T4, however, Feature fusion achieves a slightly higher success rate (87.5\% versus 85.0\%).

\begin{figure}[t]
\centering
\definecolor{camorange}{RGB}{219,122,31}
\definecolor{camgray}{RGB}{118,123,128}
\definecolor{camtrack}{RGB}{233,233,233}
\providecommand{\cambar}[2]{\makebox[\linewidth][l]{%
\textcolor{#1}{\rule[0pt]{\dimexpr#2\linewidth/40\relax}{2.8pt}}%
\textcolor{camtrack}{\rule[0pt]{\dimexpr\linewidth-#2\linewidth/40\relax}{2.8pt}}}}
\providecommand{\camcol}[6]{\begingroup\setlength{\fboxsep}{0pt}\setlength{\fboxrule}{#6}%
\begin{minipage}[t]{\camimg}\centering
\parbox[c][1.25\baselineskip][c]{\linewidth}{\centering\footnotesize\sffamily #1}\\[2pt]
\fcolorbox{#2}{white}{\includegraphics[width=\dimexpr\linewidth-2\fboxrule\relax]{#3}}\\[2.5pt]
{\footnotesize\sffamily\bfseries\textcolor{#2}{#4}}\\[2.5pt]
\cambar{#2}{#5}\end{minipage}\endgroup}
\includegraphics[width=\textwidth]{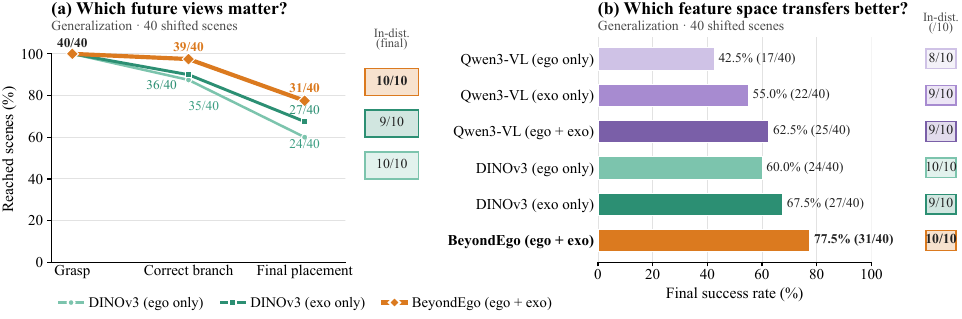}\\[4pt]
{\centering\small\sffamily\bfseries\color{yinwangdark}(c) Exocentric camera placement on T1\par}\vspace{2pt}
\ifdefined\camimg\else\newlength{\camimg}\fi%
\setlength{\camimg}{\dimexpr(\textwidth-46pt)/4\relax}%
\makebox[\textwidth][c]{%
\begin{minipage}[t]{\camimg}\centering
\parbox[c][1.25\baselineskip][c]{\linewidth}{\centering\footnotesize\sffamily Egocentric (shared)}\\[2pt]
{\setlength{\fboxsep}{0pt}\setlength{\fboxrule}{0.6pt}%
\fcolorbox{camgray}{white}{\includegraphics[width=\dimexpr\linewidth-1.2pt\relax]{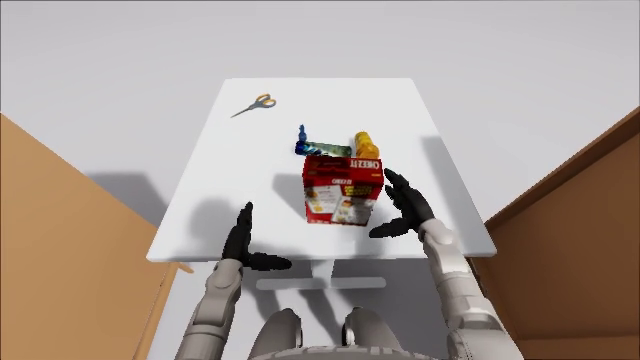}}}\\[2.5pt]
{\footnotesize\sffamily\color{camgray}identical for all}
\end{minipage}\hspace{6pt}%
\begin{minipage}[t]{18pt}\centering
\parbox[c][1.25\baselineskip][c]{\linewidth}{}\\[2pt]
\parbox[c][\dimexpr\camimg*9/16\relax][c]{\linewidth}{\centering\Large$+$}
\end{minipage}\hspace{6pt}%
\camcol{Left}{camgray}{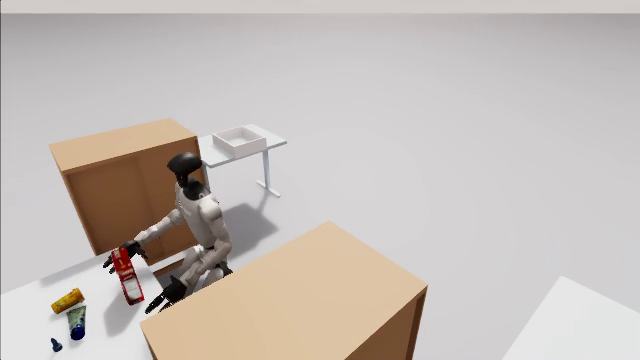}{60.0\%}{24}{0.6pt}\hspace{8pt}%
\camcol{\bfseries Central (BeyondEgo)}{camorange}{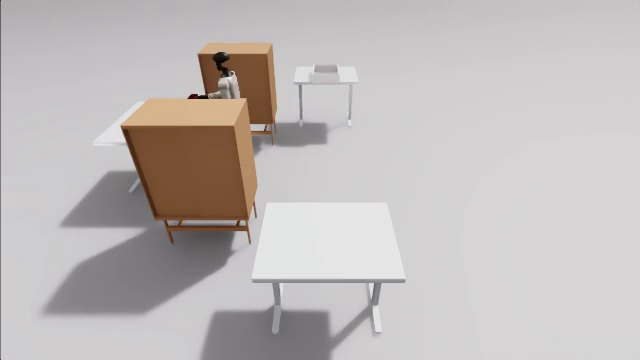}{77.5\%}{31}{1.2pt}\hspace{8pt}%
\camcol{Right}{camgray}{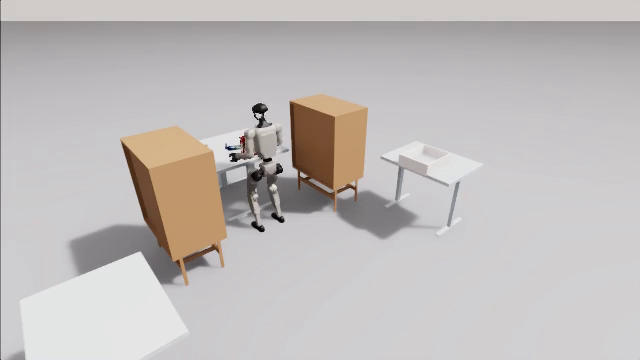}{72.5\%}{29}{0.6pt}}\\[3pt]
{\scriptsize\sffamily\color{camgray}Generalization success rate (counts\,/40)}
\caption{\textbf{T1 ablations: prediction targets, execution behavior, and exocentric camera placement.} (a) Cumulative completion of grasping, correct branch selection, and placement across 40 generalization scenes under egocentric-only, exocentric-only, and dual-view DINOv3 supervision. (b) Generalization success rates for different combinations of feature targets and supervised views. (c) All three configurations share the egocentric view (left) and add one exocentric camera ($+$) at different placements; the central one (orange) is \Method's deployed view, and values below give generalization success over the 40 shifted scenes.}
\label{fig:analysis}
\label{fig:camera}
\end{figure}

\paragraph{Random seeds}
We independently train BeyondEgo on T1 with three random seeds and evaluate each model five times on the same 40 generalization scenes (15 evaluations, 600 shifted episodes in total; \Cref{tab:seeds}). All three models achieve 100\% (10/10) in-distribution success. Under shift, all 15 evaluations fall within 75.0\%--82.5\% (30/40--33/40), averaging 78.2\% $\pm$ 1.0\% across seeds. These results indicate consistent T1 performance across the tested training seeds.

\subsection{Ablation Studies}
\label{sec:ablations}

\paragraph{Spatial fusion}\label{sec:where}
Feature fusion improves success rates over VLM fusion by 10.0 and 5.5 percentage points in the in-distribution and generalization evaluations, respectively (\Cref{fig:main}). It achieves higher success rates on all five tasks under generalization and matches or exceeds VLM fusion in distribution. We hypothesize that the additional pre-training in VLM fusion may alter the original VLM feature distribution, whereas Feature fusion supplements exocentric information through a separate DINOv3 branch, helping maintain compatibility between the original visual features and the pre-trained action expert. Given its higher success rates and absence of additional backbone pre-training, we adopt Feature fusion for the temporal extension.

\paragraph{Foresight supervision}
By adding dual-view foresight to Feature fusion, BeyondEgo improves mean success rates by 8.0 and 10.0 percentage points in the in-distribution and generalization evaluations, respectively. Future supervision may help the policy anticipate changes in the scene and the robot's position, leading to better action decisions. Both policies receive the same current observations at deployment; BeyondEgo removes the foresight expert and retains the trained adapter as its only additional component.

\paragraph{Supervised views}
For the T1 task, dual-view foresight with DINOv3 targets achieves a higher success rate than either single-view variant or the Feature fusion baseline without foresight~(Figures~\ref{fig:main} and~\ref{fig:analysis}a). This advantage may result from complementary future information: the egocentric view captures local scene changes, while the exocentric view captures changes in the robot's position within the environment. Predicting both types of future change may help the policy coordinate local manipulation with whole-body motion and improve task completion.

\paragraph{Target feature space}\label{sec:what}
We compare future feature targets from two frozen teacher models, DINOv3 and Qwen3-VL, on T1. For each target space, we use egocentric-only, exocentric-only, and dual-view supervision, yielding six variants. All variants receive current dual-view observations at deployment and differ only in the future supervision used during training.
In the generalization evaluation, DINOv3 targets yield higher task success rates than Qwen3-VL targets under all three supervision settings (\Cref{fig:analysis}b).
We argue that DINOv3 features may better preserve spatial relationships between the robot and its environment, providing more useful future supervision for turning and route selection. These results support using DINOv3 features as prediction targets for our method.

\paragraph{Exocentric camera placement}
We compare left, right, and central exocentric camera placements on T1, training and evaluating a separate model with each placement (\Cref{fig:camera}c). Neither side camera covers the full task scene, and the left view is more heavily occluded; the central placement covers the entire task area. All three configurations achieve 100\% (10/10) in-distribution success, while generalization performance varies: the central camera achieves the highest success rate of 77.5\% (31/40), followed by the right camera at 72.5\% (29/40) and the left camera at 60.0\% (24/40). Broader scene coverage may provide more complete spatial context for route selection and whole-body motion. These results support selecting exocentric camera positions that cover the full task scene with minimal occlusion.

\section{Conclusion}
\label{sec:conclusion}

We introduce BeyondEgo, which extends the perception of hierarchical humanoid VLAs in both space and time. We compare ways of integrating the exocentric view and use dual-view future supervision to improve the observations fed to the action expert. To evaluate perception beyond the egocentric view, we also introduce HELM, a closed-loop humanoid loco-manipulation benchmark with synchronized ego-exo views and controlled scene shifts. Experiments show that directly integrating exocentric information into the action expert yields better performance, while the temporal extension with an adapter and foresight supervision further improves overall task success and generalization to scene variations. BeyondEgo uses future information as training supervision; at deployment, the foresight expert is removed, and only the lightweight adapter is retained. More broadly, our results suggest that privileged signals available only during training, such as future observations from both views, can yield deployable gains through representation-level supervision. We hope this combination of exocentric perception and dual-view future supervision can inform loco-manipulation across robots and sensors.

\paragraph{Limitations}
BeyondEgo requires external camera coverage and synchronized dual-view images; large-scale movement may limit exocentric observations, and dual-view demonstrations increase data-collection cost. Future work could explore multi-camera coordination, more flexible placement, and foresight horizons beyond one action chunk.

\setlength{\bibsep}{2.6pt plus 0.8pt minus 0.5pt}
\bibliographystyle{unsrtnat}
\bibliography{main}

@inproceedings{wei2026psi0,
  title     = {{$\Psi_0$}: An Open Foundation Model Towards Universal Humanoid Loco-Manipulation},
  author    = {Songlin Wei and Hongyi Jing and Boqian Li and Zhenyu Zhao and Jiageng Mao and Zhenhao Ni and Sicheng He and Sheng Zang and Xiawei Liu and Kaidi Kang and Jie Liu and Weiduo Yuan and Marco Pavone and Di Huang and Yue Wang},
  booktitle = {Robotics: Science and Systems (RSS)},
  year      = {2026}
}

@article{bjorck2025gr00t,
  title   = {{GR00T N1}: An Open Foundation Model for Generalist Humanoid Robots},
  author  = {{NVIDIA} and Johan Bjorck and Fernando Casta{\~n}eda and Nikita Cherniadev and Xingye Da and Runyu Ding and Linxi Fan and Yu Fang and others},
  journal = {arXiv preprint arXiv:2503.14734},
  year    = {2025}
}

@inproceedings{black2024pi0,
  title     = {{$\pi_0$}: A Vision-Language-Action Flow Model for General Robot Control},
  author    = {Kevin Black and Noah Brown and Danny Driess and Adnan Esmail and Michael Robert Equi and Chelsea Finn and Niccolo Fusai and Lachy Groom and others},
  booktitle = {Robotics: Science and Systems (RSS)},
  year      = {2025}
}

@inproceedings{intelligence2025pi05,
  title     = {{$\pi_{0.5}$}: a Vision-Language-Action Model with Open-World Generalization},
  author    = {Kevin Black and Noah Brown and James Darpinian and Karan Dhabalia and Danny Driess and Adnan Esmail and Michael Robert Equi and Chelsea Finn and others},
  booktitle = {Conference on Robot Learning (CoRL)},
  year      = {2025}
}

@inproceedings{kim2024openvla,
  title     = {{OpenVLA}: An Open-Source Vision-Language-Action Model},
  author    = {Moo Jin Kim and Karl Pertsch and Siddharth Karamcheti and Ted Xiao and Ashwin Balakrishna and Suraj Nair and Rafael Rafailov and Ethan P. Foster and Pannag R. Sanketi and Quan Vuong and Thomas Kollar and Benjamin Burchfiel and Russ Tedrake and Dorsa Sadigh and Sergey Levine and Percy Liang and Chelsea Finn},
  booktitle = {Conference on Robot Learning (CoRL)},
  year      = {2024}
}

@inproceedings{chi2023diffusion,
  title     = {Diffusion Policy: Visuomotor Policy Learning via Action Diffusion},
  author    = {Cheng Chi and Siyuan Feng and Yilun Du and Zhenjia Xu and Eric Cousineau and Benjamin C. M. Burchfiel and Shuran Song},
  booktitle = {Robotics: Science and Systems (RSS)},
  year      = {2023}
}

@inproceedings{zhao2023act,
  title     = {Learning Fine-Grained Bimanual Manipulation with Low-Cost Hardware},
  author    = {Tony Z. Zhao and Vikash Kumar and Sergey Levine and Chelsea Finn},
  booktitle = {Robotics: Science and Systems (RSS)},
  year      = {2023}
}

@inproceedings{lipman2023flow,
  title     = {Flow Matching for Generative Modeling},
  author    = {Yaron Lipman and Ricky T. Q. Chen and Heli Ben-Hamu and Maximilian Nickel and Matthew Le},
  booktitle = {International Conference on Learning Representations (ICLR)},
  year      = {2023}
}

@article{bai2025qwen3vl,
  title   = {{Qwen3-VL} Technical Report},
  author  = {Shuai Bai and Yuxuan Cai and Ruizhe Chen and Keqin Chen and Xionghui Chen and Zesen Cheng and Lianghao Deng and Wei Ding and others},
  journal = {arXiv preprint arXiv:2511.21631},
  year    = {2025}
}

@article{oquab2024dinov2,
  title   = {{DINOv2}: Learning Robust Visual Features without Supervision},
  author  = {Maxime Oquab and Timoth{\'e}e Darcet and Th{\'e}o Moutakanni and Huy V. Vo and Marc Szafraniec and Vasil Khalidov and Pierre Fernandez and Daniel Haziza and others},
  journal = {Transactions on Machine Learning Research (TMLR)},
  year    = {2024}
}

@article{simeoni2025dinov3,
  title   = {{DINOv3}},
  author  = {Oriane Sim{\'e}oni and Huy V. Vo and Maximilian Seitzer and Federico Baldassarre and Maxime Oquab and Cijo Jose and Vasil Khalidov and Marc Szafraniec and others},
  journal = {Transactions on Machine Learning Research (TMLR)},
  year    = {2026}
}

@inproceedings{black2025rtc,
  title     = {Real-Time Execution of Action Chunking Flow Policies},
  author    = {Kevin Black and Manuel Y. Galliker and Sergey Levine},
  booktitle = {Advances in Neural Information Processing Systems (NeurIPS)},
  year      = {2025}
}

@article{black2025trainingrtc,
  title   = {Training-Time Action Conditioning for Efficient Real-Time Chunking},
  author  = {Kevin Black and Allen Z. Ren and Michael Equi and Sergey Levine},
  journal = {arXiv preprint arXiv:2512.05964},
  year    = {2025}
}

@inproceedings{pertsch2025fast,
  title     = {{FAST}: Efficient Action Tokenization for Vision-Language-Action Models},
  author    = {Karl Pertsch and Kyle Stachowicz and Brian Ichter and Danny Driess and Suraj Nair and Quan Vuong and Oier Mees and Chelsea Finn and Sergey Levine},
  booktitle = {Robotics: Science and Systems (RSS)},
  year      = {2025}
}

@inproceedings{li2025amo,
  title     = {{AMO}: Adaptive Motion Optimization for Hyper-Dexterous Humanoid Whole-Body Control},
  author    = {Jialong Li and Xuxin Cheng and Tianshu Huang and Shiqi Yang and Ri-Zhao Qiu and Xiaolong Wang},
  booktitle = {Robotics: Science and Systems (RSS)},
  year      = {2025}
}

@inproceedings{fu2024humanplus,
  title     = {{HumanPlus}: Humanoid Shadowing and Imitation from Humans},
  author    = {Zipeng Fu and Qingqing Zhao and Qi Wu and Gordon Wetzstein and Chelsea Finn},
  booktitle = {Conference on Robot Learning (CoRL)},
  year      = {2024}
}

@inproceedings{he2024omnih2o,
  title     = {{OmniH2O}: Universal and Dexterous Human-to-Humanoid Whole-Body Teleoperation and Learning},
  author    = {Tairan He and Zhengyi Luo and Xialin He and Wenli Xiao and Chong Zhang and Weinan Zhang and Kris M. Kitani and Changliu Liu and Guanya Shi},
  booktitle = {Conference on Robot Learning (CoRL)},
  year      = {2024}
}

@inproceedings{jiang2025wholebodyvla,
  title     = {{WholeBodyVLA}: Towards Unified Latent {VLA} for Whole-body Loco-manipulation Control},
  author    = {Haoran Jiang and Jin Chen and Qingwen Bu and Li Chen and Modi Shi and Yanjie Zhang and Delong Li and Chuanzhe Suo and Chuang Wang and Zhihui Peng and Hongyang Li},
  booktitle = {International Conference on Learning Representations (ICLR)},
  year      = {2026}
}

@article{ding2025humanoidvla,
  title   = {{Humanoid-VLA}: Towards Universal Humanoid Control with Visual Integration},
  author  = {Pengxiang Ding and Jianfei Ma and Xinyang Tong and Binghong Zou and Xinxin Luo and Yiguo Fan and Ting Wang and Hongchao Lu and Panzhong Mo and Jinxin Liu and Yuefan Wang and Huaicheng Zhou and Wenshuo Feng and Jiacheng Liu and Siteng Huang and Donglin Wang},
  journal = {arXiv preprint arXiv:2502.14795},
  year    = {2025}
}

@article{xue2025leverb,
  title   = {{LeVERB}: Humanoid Whole-Body Control with Latent Vision-Language Instruction},
  author  = {Haoru Xue and Xiaoyu Huang and Dantong Niu and Qiayuan Liao and Thomas Kragerud and Jan Tommy Gravdahl and Xue Bin Peng and Guanya Shi and Trevor Darrell and Koushil Sreenath and Shankar Sastry},
  journal = {arXiv preprint arXiv:2506.13751},
  year    = {2025}
}

@inproceedings{grauman2024egoexo4d,
  title     = {{Ego-Exo4D}: Understanding Skilled Human Activity from First- and Third-Person Perspectives},
  author    = {Kristen Grauman and Andrew Westbury and Lorenzo Torresani and Kris Kitani and Jitendra Malik and Triantafyllos Afouras and Kumar Ashutosh and Vijay Baiyya and others},
  booktitle = {Proceedings of the IEEE/CVF Conference on Computer Vision and Pattern Recognition (CVPR)},
  year      = {2024}
}

@inproceedings{seo2023mvmwm,
  title     = {Multi-View Masked World Models for Visual Robotic Manipulation},
  author    = {Younggyo Seo and Junsu Kim and Stephen James and Kimin Lee and Jinwoo Shin and Pieter Abbeel},
  booktitle = {International Conference on Machine Learning (ICML)},
  year      = {2023}
}

@article{vapnik2009lupi,
  title   = {A New Learning Paradigm: Learning Using Privileged Information},
  author  = {Vladimir Vapnik and Akshay Vashist},
  journal = {Neural Networks},
  volume  = {22},
  number  = {5--6},
  pages   = {544--557},
  year    = {2009}
}

@inproceedings{schwarzer2021spr,
  title     = {Data-Efficient Reinforcement Learning with Self-Predictive Representations},
  author    = {Max Schwarzer and Ankesh Anand and Rishab Goel and R. Devon Hjelm and Aaron Courville and Philip Bachman},
  booktitle = {International Conference on Learning Representations (ICLR)},
  year      = {2021}
}

@article{assran2025vjepa2,
  title   = {{V-JEPA 2}: Self-Supervised Video Models Enable Understanding, Prediction and Planning},
  author  = {Mido Assran and Adrien Bardes and David Fan and Quentin Garrido and Russell Howes and Mojtaba Komeili and Matthew Muckley and Ammar Rizvi and others},
  journal = {arXiv preprint arXiv:2506.09985},
  year    = {2025}
}

@inproceedings{zhou2025dinowm,
  title     = {{DINO-WM}: World Models on Pre-trained Visual Features enable Zero-shot Planning},
  author    = {Gaoyue Zhou and Hengkai Pan and Yann LeCun and Lerrel Pinto},
  booktitle = {International Conference on Machine Learning (ICML)},
  year      = {2025}
}

@inproceedings{zheng2025flare,
  title     = {{FLARE}: Robot Learning with Implicit World Modeling},
  author    = {Ruijie Zheng and Jing Wang and Scott Reed and Johan Bjorck and Yu Fang and Fengyuan Hu and Joel Jang and Kaushil Kundalia and others},
  booktitle = {Conference on Robot Learning (CoRL)},
  year      = {2025}
}

@inproceedings{tian2025seer,
  title     = {Predictive Inverse Dynamics Models are Scalable Learners for Robotic Manipulation},
  author    = {Yang Tian and Sizhe Yang and Jia Zeng and Ping Wang and Dahua Lin and Hao Dong and Jiangmiao Pang},
  booktitle = {International Conference on Learning Representations (ICLR)},
  year      = {2025}
}

@inproceedings{hu2025vpp,
  title     = {Video Prediction Policy: A Generalist Robot Policy with Predictive Visual Representations},
  author    = {Yucheng Hu and Yanjiang Guo and Pengchao Wang and Xiaoyu Chen and Yen-Jen Wang and Jianke Zhang and Koushil Sreenath and Chaochao Lu and Jianyu Chen},
  booktitle = {International Conference on Machine Learning (ICML)},
  year      = {2025}
}

@inproceedings{wu2024gr1,
  title     = {Unleashing Large-Scale Video Generative Pre-training for Visual Robot Manipulation},
  author    = {Hongtao Wu and Ya Jing and Chilam Cheang and Guangzeng Chen and Jiafeng Xu and Xinghang Li and Minghuan Liu and Hang Li and Tao Kong},
  booktitle = {International Conference on Learning Representations (ICLR)},
  year      = {2024}
}

@inproceedings{zhao2025cotvla,
  title     = {{CoT-VLA}: Visual Chain-of-Thought Reasoning for Vision-Language-Action Models},
  author    = {Qingqing Zhao and Yao Lu and Moo Jin Kim and Zipeng Fu and Zhuoyang Zhang and Yecheng Wu and Zhaoshuo Li and Qianli Ma and Song Han and Chelsea Finn and Ankur Handa and Tsung-Yi Lin and Gordon Wetzstein and Ming-Yu Liu and Donglai Xiang},
  booktitle = {Proceedings of the Computer Vision and Pattern Recognition Conference (CVPR)},
  year      = {2025}
}

@inproceedings{zhang2025upvla,
  title     = {{UP-VLA}: A Unified Understanding and Prediction Model for Embodied Agent},
  author    = {Jianke Zhang and Yanjiang Guo and Yucheng Hu and Xiaoyu Chen and Xiang Zhu and Jianyu Chen},
  booktitle = {International Conference on Machine Learning (ICML)},
  year      = {2025}
}

@inproceedings{zhang2025dreamvla,
  title     = {{DreamVLA}: A Vision-Language-Action Model Dreamed with Comprehensive World Knowledge},
  author    = {Wenyao Zhang and Hongsi Liu and Zekun Qi and Yunnan Wang and Xinqiang Yu and Jiazhao Zhang and Runpei Dong and Jiawei He and He Wang and Zhizheng Zhang and Li Yi and Wenjun Zeng and Xin Jin},
  booktitle = {Advances in Neural Information Processing Systems (NeurIPS)},
  year      = {2025}
}

@article{cen2025worldvla,
  title   = {{WorldVLA}: Towards Autoregressive Action World Model},
  author  = {Jun Cen and Chaohui Yu and Hangjie Yuan and Yuming Jiang and Siteng Huang and Jiayan Guo and Xin Li and Yibing Song and Hao Luo and Fan Wang and Deli Zhao and Hao Chen},
  journal = {arXiv preprint arXiv:2506.21539},
  year    = {2025}
}

@inproceedings{du2023unipi,
  title     = {Learning Universal Policies via Text-Guided Video Generation},
  author    = {Yilun Du and Sherry Yang and Bo Dai and Hanjun Dai and Ofir Nachum and Josh Tenenbaum and Dale Schuurmans and Pieter Abbeel},
  booktitle = {Advances in Neural Information Processing Systems (NeurIPS)},
  year      = {2023}
}

@inproceedings{yu2025repa,
  title     = {Representation Alignment for Generation: Training Diffusion Transformers Is Easier Than You Think},
  author    = {Sihyun Yu and Sangkyung Kwak and Huiwon Jang and Jongheon Jeong and Jonathan Huang and Jinwoo Shin and Saining Xie},
  booktitle = {International Conference on Learning Representations (ICLR)},
  year      = {2025}
}

@inproceedings{nair2022r3m,
  title     = {{R3M}: A Universal Visual Representation for Robot Manipulation},
  author    = {Suraj Nair and Aravind Rajeswaran and Vikash Kumar and Chelsea Finn and Abhinav Gupta},
  booktitle = {Conference on Robot Learning (CoRL)},
  year      = {2022}
}

@inproceedings{majumdar2023vc1,
  title     = {Where are we in the search for an Artificial Visual Cortex for Embodied Intelligence?},
  author    = {Arjun Majumdar and Karmesh Yadav and Sergio Arnaud and Yecheng Jason Ma and Claire Chen and Sneha Silwal and Aryan Jain and Vincent-Pierre Berges and Tingfan Wu and Jay Vakil and Pieter Abbeel and Jitendra Malik and Dhruv Batra and Yixin Lin and Oleksandr Maksymets and Aravind Rajeswaran and Franziska Meier},
  booktitle = {Advances in Neural Information Processing Systems (NeurIPS)},
  year      = {2023}
}

@inproceedings{simple_sim,
  title     = {{SIMPLE}: Simulation-Based Policy Learning and Evaluation for Humanoid Loco-manipulation},
  author    = {Songlin Wei and Zhenhao Ni and Jie Liu and Zhenyu Zhao and Junjie Ye and Hongyi Jing and Junkai Xia and Xiawei Liu and Michael Leong and Liang Heng and Di Huang and Yue Wang},
  booktitle = {Conference on Robot Learning (CoRL)},
  year      = {2026},
  note      = {Code: \url{https://github.com/physical-superintelligence-lab/SIMPLE}}
}

@inproceedings{brohan2023rt2,
  title     = {{RT-2}: Vision-Language-Action Models Transfer Web Knowledge to Robotic Control},
  author    = {Brohan, Anthony and Brown, Noah and Carbajal, Justice and Chebotar, Yevgen and Chen, Xi and Choromanski, Krzysztof and Ding, Tianli and Driess, Danny and others},
  booktitle = {Conference on Robot Learning (CoRL)},
  year      = {2023}
}

@article{cheng2024navila,
  title   = {{NaVILA}: Legged Robot Vision-Language-Action Model for Navigation},
  author  = {Cheng, An-Chieh and Ji, Yandong and Yang, Zhaojing and Gongye, Zaitian and Zou, Xueyan and Kautz, Jan and B{\i}y{\i}k, Erdem and Yin, Hongxu and others},
  journal = {arXiv preprint arXiv:2412.04453},
  year    = {2024}
}

@article{yang2025egovla,
  title   = {{EgoVLA}: Learning Vision-Language-Action Models from Egocentric Human Videos},
  author  = {Yang, Ruihan and Yu, Qinxi and Wu, Yecheng and Yan, Rui and Li, Borui and Cheng, An-Chieh and Zou, Xueyan and Fang, Yunhao and others},
  journal = {arXiv preprint arXiv:2507.12440},
  year    = {2025}
}

@article{ben2025homie,
  title   = {{HOMIE}: Humanoid Loco-Manipulation with Isomorphic Exoskeleton Cockpit},
  author  = {Ben, Qingwei and Jia, Feiyu and Zeng, Jia and Dong, Junting and Lin, Dahua and Pang, Jiangmiao},
  journal = {arXiv preprint arXiv:2502.13013},
  year    = {2025}
}

@inproceedings{hsu2022seeing,
  title     = {Vision-Based Manipulators Need to Also See from Their Hands},
  author    = {Hsu, Kyle and Kim, Moo Jin and Rafailov, Rafael and Wu, Jiajun and Finn, Chelsea},
  booktitle = {International Conference on Learning Representations (ICLR)},
  year      = {2022}
}

@inproceedings{sermanet2018tcn,
  title     = {Time-Contrastive Networks: Self-Supervised Learning from Video},
  author    = {Sermanet, Pierre and Lynch, Corey and Chebotar, Yevgen and Hsu, Jasmine and Jang, Eric and Schaal, Stefan and Levine, Sergey},
  booktitle = {IEEE International Conference on Robotics and Automation (ICRA)},
  year      = {2018}
}

@inproceedings{sigurdsson2018actor,
  title     = {Actor and Observer: Joint Modeling of First and Third-Person Videos},
  author    = {Sigurdsson, Gunnar A. and Gupta, Abhinav and Schmid, Cordelia and Farhadi, Ali and Alahari, Karteek},
  booktitle = {Proceedings of the IEEE Conference on Computer Vision and Pattern Recognition (CVPR)},
  year      = {2018}
}

@inproceedings{huang2024egoexolearn,
  title     = {{EgoExoLearn}: A Dataset for Bridging Asynchronous Ego- and Exo-centric View of Procedural Activities in Real World},
  author    = {Huang, Yifei and Chen, Guo and Xu, Jilan and Zhang, Mingfang and Yang, Lijin and Pei, Baoqi and Zhang, Hongjie and Dong, Lu and others},
  booktitle = {Proceedings of the IEEE/CVF Conference on Computer Vision and Pattern Recognition (CVPR)},
  year      = {2024}
}

@article{li2025pointvla,
  title   = {{PointVLA}: Injecting the {3D} World into Vision-Language-Action Models},
  author  = {Li, Chengmeng and Wen, Junjie and Peng, Yan and Peng, Yaxin and Feng, Feifei and Zhu, Yichen},
  journal = {arXiv preprint arXiv:2503.07511},
  year    = {2025}
}

@article{sun2025geovla,
  title   = {{GeoVLA}: Empowering {3D} Representations in Vision-Language-Action Models},
  author  = {Sun, Lin and Xie, Bin and Liu, Yingfei and Shi, Hao and Wang, Tiancai and Cao, Jiale},
  journal = {arXiv preprint arXiv:2508.09071},
  year    = {2025}
}

@article{hafner2023dreamerv3,
  title   = {Mastering Diverse Domains through World Models},
  author  = {Hafner, Danijar and Pasukonis, Jurgis and Ba, Jimmy and Lillicrap, Timothy},
  journal = {arXiv preprint arXiv:2301.04104},
  year    = {2023}
}

@inproceedings{bar2025nwm,
  title     = {Navigation World Models},
  author    = {Bar, Amir and Zhou, Gaoyue and Tran, Danny and Darrell, Trevor and LeCun, Yann},
  booktitle = {Proceedings of the IEEE/CVF Conference on Computer Vision and Pattern Recognition (CVPR)},
  year      = {2025}
}

@article{li2025spatialforcing,
  title   = {Spatial Forcing: Implicit Spatial Representation Alignment for Vision-Language-Action Model},
  author  = {Li, Fuhao and Song, Wenxuan and Zhao, Han and Wang, Jingbo and Ding, Pengxiang and Wang, Donglin and Zeng, Long and Li, Haoang},
  journal = {arXiv preprint arXiv:2510.12276},
  year    = {2025}
}

@article{song2025reconvla,
  title   = {{ReconVLA}: Reconstructive Vision-Language-Action Model as Effective Robot Perceiver},
  author  = {Song, Wenxuan and Zhou, Ziyang and Zhao, Han and Chen, Jiayi and Ding, Pengxiang and Yan, Haodong and Huang, Yuxin and Tang, Feilong and others},
  journal = {arXiv preprint arXiv:2508.10333},
  year    = {2025}
}

@inproceedings{chen2019lbc,
  title     = {Learning by Cheating},
  author    = {Chen, Dian and Zhou, Brady and Koltun, Vladlen and Kr{\"a}henb{\"u}hl, Philipp},
  booktitle = {Conference on Robot Learning (CoRL)},
  year      = {2019}
}

@article{lee2020quadruped,
  title   = {Learning Quadrupedal Locomotion over Challenging Terrain},
  author  = {Lee, Joonho and Hwangbo, Jemin and Wellhausen, Lorenz and Koltun, Vladlen and Hutter, Marco},
  journal = {Science Robotics},
  volume  = {5},
  number  = {47},
  pages   = {eabc5986},
  year    = {2020}
}

@inproceedings{kumar2021rma,
  title     = {{RMA}: Rapid Motor Adaptation for Legged Robots},
  author    = {Kumar, Ashish and Fu, Zipeng and Pathak, Deepak and Malik, Jitendra},
  booktitle = {Robotics: Science and Systems (RSS)},
  year      = {2021}
}

@inproceedings{pinto2018asymmetric,
  title     = {Asymmetric Actor Critic for Image-Based Robot Learning},
  author    = {Pinto, Lerrel and Andrychowicz, Marcin and Welinder, Peter and Zaremba, Wojciech and Abbeel, Pieter},
  booktitle = {Robotics: Science and Systems (RSS)},
  year      = {2018}
}

@inproceedings{elbanani2024probe3d,
  title     = {Probing the {3D} Awareness of Visual Foundation Models},
  author    = {El Banani, Mohamed and Raj, Amit and Maninis, Kevis-Kokitsi and Kar, Abhishek and Li, Yuanzhen and Rubinstein, Michael and Sun, Deqing and Guibas, Leonidas and others},
  booktitle = {Proceedings of the IEEE/CVF Conference on Computer Vision and Pattern Recognition (CVPR)},
  year      = {2024}
}

@inproceedings{karamcheti2024prismatic,
  title     = {Prismatic {VLMs}: Investigating the Design Space of Visually-Conditioned Language Models},
  author    = {Karamcheti, Siddharth and Nair, Suraj and Balakrishna, Ashwin and Liang, Percy and Kollar, Thomas and Sadigh, Dorsa},
  booktitle = {International Conference on Machine Learning (ICML)},
  year      = {2024}
}

@inproceedings{pumacay2024colosseum,
  title     = {{THE COLOSSEUM}: A Benchmark for Evaluating Generalization for Robotic Manipulation},
  author    = {Pumacay, Wilbert and Singh, Ishika and Duan, Jiafei and Krishna, Ranjay and Thomason, Jesse and Fox, Dieter},
  booktitle = {Robotics: Science and Systems (RSS)},
  year      = {2024}
}

@inproceedings{li2024simpler,
  title     = {Evaluating Real-World Robot Manipulation Policies in Simulation},
  author    = {Li, Xuanlin and Hsu, Kyle and Gu, Jiayuan and Pertsch, Karl and Mees, Oier and Walke, Homer Rich and Fu, Chuyuan and Lunawat, Ishikaa and others},
  booktitle = {Conference on Robot Learning (CoRL)},
  year      = {2024}
}

@article{li2026omega0,
  title   = {{$\omega$-0: A Latent Predictive World Action Model for Concurrent Humanoid Loco-Manipulation}},
  author  = {Zhe Li and Zhenzhe Zhang and Yangyang Wei and Wenjie Zhang and
             Xichen Yuan and Peiyuan Zhi and Gen Li and Xinying Guo and
             Fengjie Gao and Jianfei Yang and Shanghang Zhang},
  journal = {arXiv preprint arXiv:2608.06375},
  year    = {2026},
  doi     = {10.48550/arXiv.2608.06375},
  url     = {https://arxiv.org/abs/2608.06375}
}

@misc{qin2026wbwam,
  title={WB-WAM: Heterogeneous Body-Hand Pre-training for Humanoid Loco-Manipulation},
  author={Chuan Qin and Shaoting Zhu and Siyuan Luo and Siqiao Huang and Hongyu Zhao and Shanaka Baduge and Hang Zhao},
  year={2026},
  eprint={2609.34199},
  archivePrefix={arXiv},
  primaryClass={cs.RO},
  url={https://arxiv.org/abs/2609.34199},
}

@article{gao2021clip,
  title={CLIP-Adapter: Better Vision-Language Models with Feature Adapters},
  author={Gao, Peng and Geng, Shijie and Zhang, Renrui and Ma, Teli and Fang, Rongyao and Zhang, Yongfeng and Li, Hongsheng and Qiao, Yu},
  journal={arXiv preprint arXiv:2110.04544},
  year={2021}
}

@InProceedings{houlsby2019param,
  title = 	 {Parameter-Efficient Transfer Learning for {NLP}},
  author =       {Houlsby, Neil and Giurgiu, Andrei and Jastrzebski, Stanislaw and Morrone, Bruna and De Laroussilhe, Quentin and Gesmundo, Andrea and Attariyan, Mona and Gelly, Sylvain},
  booktitle = 	 {Proceedings of the 36th International Conference on Machine Learning},
  pages = 	 {2790--2799},
  year = 	 {2019},
  editor = 	 {Chaudhuri, Kamalika and Salakhutdinov, Ruslan},
  volume = 	 {97},
  series = 	 {Proceedings of Machine Learning Research},
  month = 	 {09--15 Jun},
  publisher =    {PMLR},
  url = 	 {https://proceedings.mlr.press/v97/houlsby19a.html}
}

\beginappendix

These appendices provide implementation details (\Cref{sec:supp-impl}), the data and benchmark protocol (\Cref{sec:supp-bench}), additional results (\Cref{sec:supp-results}), the verification that the foresight expert can be removed (\Cref{sec:supp-prune}), pseudo-code (\Cref{sec:supp-algo}) and further discussion (\Cref{sec:supp-discussion}).

\section{Implementation details}
\label{sec:supp-impl}

\begin{table}[!htb]
\centering
\caption{\textbf{Architecture.} Trained: updated during task fine-tuning. Deployed: part of the inference graph. The trainable action path (projection, adapter and action expert) has 497.6M parameters.}
\label{tab:supp-arch}
\small
\renewcommand{\arraystretch}{1.18}
\begin{tabularx}{\textwidth}{l Y r c c}
\toprule
\sffamily\bfseries Component &
\sffamily\bfseries Configuration &
\sffamily\bfseries Params &
\sffamily\bfseries Trained &
\sffamily\bfseries Deployed \\
\midrule
VLM (System-2) & Qwen3-VL-2B-Instruct~\cite{bai2025qwen3vl} with \PsiZero pre-training~\cite{wei2026psi0}; egocentric image at $256\times448$ and instruction & 2B & -- & \checkmark \\
Exocentric encoder $\Dino$ & DINOv3 ViT-L/16~\cite{simeoni2025dinov3} at $224\times384$; class token and $2\times2$ average-pooled $7\times12$ patch tokens (85 tokens of width 1024); register tokens dropped & 0.3B & -- & \checkmark \\
Exocentric projection $P_\psi$ & $1024\rightarrow2048$ & -- & \checkmark & \checkmark \\
Ego adapter & $\widetilde Z^{\ego}=Z^{\ego}+W_\alpha\LN(Z^{\ego})$, $W_\alpha\in\mathbb R^{2048\times2048}$ zero-initialized & 4.2M & \checkmark & \checkmark \\
Action expert (System-1) & \PsiZero flow-matching MM-DiT, initialized from \PsiZero pre-training; chunk $30\times36$ & ${\approx}$0.5B & \checkmark & \checkmark \\
Foresight expert $F_\omega$ & Width 512. Memory: ego and exo tokens (LN, linear $2048\rightarrow512$), state token (two-layer MLP), learned null token. Per view: 85 learned queries, 2 blocks of query self-attention, cross-attention to the memory and FFN (8 heads, FFN width 2048, GELU, pre-LN, dropout 0.1), linear $512\rightarrow1024$ & 19.2M & \checkmark & -- \\
Foresight teacher & The frozen $\Dino$ applied to $I^{\ego}_{t+\Delta t}$ and $I^{\exo}_{t+\Delta t}$ with the same tokenization; stop-gradient & shared & -- & -- \\
System-0 & AMO whole-body controller~\cite{li2025amo} & -- & -- & \checkmark \\
\bottomrule
\end{tabularx}
\end{table}

\paragraph{Architecture}
\Cref{tab:supp-arch} lists all components.
The egocentric VLM tokens and the projected exocentric tokens form the observation sequence of the action expert (\Cref{eq:act}); when the exocentric view is dropped during training, its tokens are masked out of the attention of both the action expert and the foresight expert $F_\omega$, and the null token of $F_\omega$ keeps its memory non-empty.

\paragraph{Action and state}
Each of the 30 commands in a chunk has 36 dimensions: 14 hand joint targets, 14 arm joint targets, torso roll, pitch and yaw, torso height, forward, lateral and yaw velocity, and a target heading.
The last eight dimensions are tracked by the AMO controller~\cite{li2025amo}, which outputs the leg and waist joint targets.
Actions and states are normalized with statistics of the training data.

\begin{table}[!htb]
\centering
\caption{\textbf{Training and evaluation hyperparameters}, shared by all methods unless noted.}
\label{tab:supp-hparams}
\small
\renewcommand{\arraystretch}{1.16}
\begin{tabularx}{\linewidth}{l Y}
\toprule
\sffamily\bfseries Parameter & \sffamily\bfseries Value \\
\midrule
Optimizer & AdamW, $\beta=(0.95,0.999)$, weight decay $10^{-6}$ \\
Learning rate & $10^{-4}$, 1000 warm-up steps, cosine decay \\
Training steps / batch & 40{,}000 / 128 (16 per GPU, 8 B200 GPUs) \\
Flow matching & 1000 discrete flow times in training; 10 Euler steps at test time \\
Action chunk & $H=30$ commands at 50\,Hz (0.6\,s) \\
Real-time chunking & training: random clean prefix of up to 8 commands~\cite{black2025trainingrtc}; test: execute 15 commands per chunk, inference delay 7~\cite{black2025rtc} \\
Exocentric dropout & 0.1 (Feature fusion and \Method) \\
Foresight (\Method) & $\lambda=1$, linear ramp over the first 10\% of steps; $\Delta t=0.6$\,s (30 frames); cosine loss \\
Training time & 8--11.5\,h per task on 8 B200 GPUs (\Method) \\
\midrule
Evaluation & SIMPLE~\cite{simple_sim}; 35\,s episode limit on all five tasks; collisions logged, not terminal \\
In-distribution suite & 10 trials per task, fixed seed, paired across methods \\
Generalization suite & 40 frozen scenes per task, 8 per shift family \\
\bottomrule
\end{tabularx}
\end{table}

\paragraph{VLM fusion}
Both views are rendered into a single VLM prompt.
The backbone is initialized from the \PsiZero pre-trained VLM and further pre-trained for about 145K steps on the 500 dual-view demonstrations of the five tasks (100 per task), i.e., the demonstrations that are also used for task fine-tuning, by next-token prediction of FAST action tokens~\cite{pertsch2025fast}.
It is then frozen; the last-layer tokens of each view are linearly projected and concatenated as the observation sequence of the action expert, which is trained with the recipe of \Cref{tab:supp-hparams}.
This variant uses no DINOv3 features.

\paragraph{Qwen3-VL teacher}
For the teacher ablation (\Cref{fig:analysis}b), the frozen Qwen3-VL backbone encodes both future frames at $224\times384$ together with the current instruction in one prompt.
For each view we take the last-layer hidden states at its $7\times12$ merged image-token positions and prepend their mean in place of a class token, which gives $85\times2048$ targets with the same layout as the DINOv3 targets; the output heads of $F_\omega$ are widened to 2048 accordingly, and all other settings are unchanged.

\section{Data and benchmark}
\label{sec:supp-bench}

\begin{table}[!htb]
\centering
\caption{\textbf{Demonstration data.} 50\,Hz, $640\times360$, egocentric head camera and fixed oblique exocentric camera. Length: mean episode duration. T3 covers five layouts.}
\label{tab:supp-data}
\small
\renewcommand{\arraystretch}{1.16}
\setlength{\tabcolsep}{7pt}
\begin{tabularx}{\columnwidth}{l Y rrr}
\toprule
\sffamily\bfseries Task &
\sffamily\bfseries Name &
\sffamily\bfseries Episodes &
\sffamily\bfseries Frames &
\sffamily\bfseries Length \\
\midrule
T1 & Back Out and Turn & 100 & \BEf{frames}{1} & 24.5\,s \\
T2 & Turn and Search & 100 & \BEf{frames}{2} & 19.0\,s \\
T3 & Choose the Right Passage & 100 & \BEf{frames}{3} & 27.4\,s \\
T4 & Sidestep and Turn & 100 & \BEf{frames}{4} & 23.7\,s \\
T5 & Sideways Passage & 100 & \BEf{frames}{5} & 34.1\,s \\
\midrule
\ywrowhighlight
\textbf{All} & & \textbf{500} & \textbf{\BEf{frames}{total}} & \textbf{25.8\,s} \\
\bottomrule
\end{tabularx}
\end{table}

\paragraph{Demonstrations}
\Cref{tab:supp-data} summarizes the data.
All 500 demonstrations are used for training; the offline metrics (\Cref{tab:supp-ablation,tab:supp-offline5}) are computed on the last 10 episodes of each task.
Every episode starts with the box on a table and ends with the box in the tray.
The simulator records four cameras; we use the head-mounted egocentric camera and one fixed oblique exocentric camera.

\paragraph{Closed-loop evaluation}
The policy server runs the deployed graph; the client runs SIMPLE~\cite{simple_sim} (Isaac Sim rendering, MuJoCo physics) with the AMO controller in the loop and no privileged inputs to the policy.
Success is evaluated by a task-specific predicate on the final state: for T1, the box must rest stably supported near the center of the tray; for T2--T5, the box must have been released into the tray.
Collisions are checked at 200\,Hz and recorded, but do not terminate a trial or disqualify success.
The T1 stage funnel of \Cref{fig:analysis}a records whether the box was grasped, whether the robot took the branch of the junction that leads to the tray, and whether the trial succeeded.

\paragraph{Generalization suite}
For each task we freeze 40 scenes, 8 per shift family: \emph{layout} (arrangements with parameters outside the training ranges, such as narrower channels and tighter obstacle gaps), \emph{asset} (unseen object and furniture meshes), \emph{material} (unseen textures), \emph{lighting} (intensity scaled by 0.55 or 1.5 and color temperature 3400\,K or 8800\,K) and \emph{combined} shifts.
Only visual bindings, materials, lighting and layout parameters change; collision geometry and scoring are untouched.
Every layout and combined scene passes a static route-feasibility check, and no scene duplicates a training episode or another scene of the suite.
Scene files are frozen and identified by SHA-256 hashes, so all methods face identical scenes.

\section{Additional results}
\label{sec:supp-results}

\begin{table}[!htb]
\centering
\caption{\textbf{Closed-loop success per task} (counts behind \Cref{fig:main}). In-distribution: successes in 10 paired trials per task. Generalization: successes in 40 frozen shifted scenes per task. Mean: success rate (\%) over all 50 and 200 episodes, with its 95\% Wilson score interval. Best per column in bold.}
\label{tab:supp-main}
\small
\renewcommand{\arraystretch}{1.18}
\setlength{\tabcolsep}{3.8pt}
\begin{tabular}{l cccccc cccccc}
\toprule
& \multicolumn{6}{c}{\sffamily\bfseries In-distribution (/10)} & \multicolumn{6}{c}{\sffamily\bfseries Generalization (/40)} \\
\cmidrule(lr){2-7}\cmidrule(lr){8-13}
\sffamily\bfseries Method &
\sffamily\bfseries T1 & \sffamily\bfseries T2 & \sffamily\bfseries T3 & \sffamily\bfseries T4 & \sffamily\bfseries T5 & \sffamily\bfseries Mean [95\% CI] &
\sffamily\bfseries T1 & \sffamily\bfseries T2 & \sffamily\bfseries T3 & \sffamily\bfseries T4 & \sffamily\bfseries T5 & \sffamily\bfseries Mean [95\% CI] \\
\midrule
\PsiZero & \BEm{id}{psi}{1} & \BEm{id}{psi}{2} & \BEm{id}{psi}{3} & \BEm{id}{psi}{4} & \BEm{id}{psi}{5} & \BEm{id}{psi}{avgb} {\scriptsize\BEm{id}{psi}{ci}} & \BEm{gen}{psi}{1} & \BEm{gen}{psi}{2} & \BEm{gen}{psi}{3} & \BEm{gen}{psi}{4} & \BEm{gen}{psi}{5} & \BEm{gen}{psi}{avgb} {\scriptsize\BEm{gen}{psi}{ci}} \\
\Stwo & \BEm{id}{s2}{1} & \BEm{id}{s2}{2} & \BEm{id}{s2}{3} & \BEm{id}{s2}{4} & \BEm{id}{s2}{5} & \BEm{id}{s2}{avgb} {\scriptsize\BEm{id}{s2}{ci}} & \BEm{gen}{s2}{1} & \BEm{gen}{s2}{2} & \BEm{gen}{s2}{3} & \BEm{gen}{s2}{4} & \BEm{gen}{s2}{5} & \BEm{gen}{s2}{avgb} {\scriptsize\BEm{gen}{s2}{ci}} \\
\Sone & \BEm{id}{s1}{1} & \BEm{id}{s1}{2} & \BEm{id}{s1}{3} & \BEm{id}{s1}{4} & \BEm{id}{s1}{5} & \BEm{id}{s1}{avgb} {\scriptsize\BEm{id}{s1}{ci}} & \BEm{gen}{s1}{1} & \BEm{gen}{s1}{2} & \BEm{gen}{s1}{3} & \BEm{gen}{s1}{4} & \BEm{gen}{s1}{5} & \BEm{gen}{s1}{avgb} {\scriptsize\BEm{gen}{s1}{ci}} \\
\midrule
\ywrowhighlight
\textbf{\Method} & \BEm{id}{full}{1} & \BEm{id}{full}{2} & \BEm{id}{full}{3} & \BEm{id}{full}{4} & \BEm{id}{full}{5} & \BEm{id}{full}{avgb} {\scriptsize\BEm{id}{full}{ci}} & \BEm{gen}{full}{1} & \BEm{gen}{full}{2} & \BEm{gen}{full}{3} & \BEm{gen}{full}{4} & \BEm{gen}{full}{5} & \BEm{gen}{full}{avgb} {\scriptsize\BEm{gen}{full}{ci}} \\
\bottomrule
\end{tabular}
\end{table}

\begin{table}[!htb]
\centering
\caption{\textbf{What to foresee} (T1; numbers behind \Cref{fig:analysis}). Cos: offline cosine similarity between predicted and target features over all frames of the last 10 T1 episodes (ego/exo). ID: in-distribution successes (/10). Generalization: shifted scenes (/40) in which the box is grasped, the robot takes the correct branch, and the box is placed successfully. \Sone has no foresight.}
\label{tab:supp-ablation}
\small
\renewcommand{\arraystretch}{1.16}
\setlength{\tabcolsep}{7pt}
\begin{tabular}{l cc ccc}
\toprule
& & & \multicolumn{3}{c}{\sffamily\bfseries Generalization (/40)} \\
\cmidrule(lr){4-6}
\sffamily\bfseries Supervised views &
\sffamily\bfseries Cos &
\sffamily\bfseries ID &
\sffamily\bfseries Grasp &
\sffamily\bfseries Branch &
\sffamily\bfseries Succ. \\
\midrule
\Sone & -- & \BEa{s1}{id}{4} & -- & -- & \BEa{s1}{gen}{4} \\
\midrule
\multicolumn{6}{l}{\textit{DINOv3 teacher}} \\
\quad ego & \BEo{dinoego}{cosego}/-- & \BEa{dinoego}{id}{4} & \BEa{dinoego}{gen}{1} & \BEa{dinoego}{gen}{2} & \BEa{dinoego}{gen}{4} \\
\quad exo & --/\BEo{dinoexo}{cosexo} & \BEa{dinoexo}{id}{4} & \BEa{dinoexo}{gen}{1} & \BEa{dinoexo}{gen}{2} & \BEa{dinoexo}{gen}{4} \\
\ywrowhighlight
\quad \textbf{both (\Method)} & \BEo{full}{cosego}/\BEo{full}{cosexo} & \BEa{full}{id}{4} & \BEa{full}{gen}{1} & \BEa{full}{gen}{2} & \BEa{full}{gen}{4} \\
\midrule
\multicolumn{6}{l}{\textit{Qwen3-VL teacher}} \\
\quad ego & \BEo{qwenego}{cosego}/-- & \BEa{qwenego}{id}{4} & \BEa{qwenego}{gen}{1} & \BEa{qwenego}{gen}{2} & \BEa{qwenego}{gen}{4} \\
\quad exo & --/\BEo{qwenexo}{cosexo} & \BEa{qwenexo}{id}{4} & \BEa{qwenexo}{gen}{1} & \BEa{qwenexo}{gen}{2} & \BEa{qwenexo}{gen}{4} \\
\quad both & \BEo{qwenboth}{cosego}/\BEo{qwenboth}{cosexo} & \BEa{qwenboth}{id}{4} & \BEa{qwenboth}{gen}{1} & \BEa{qwenboth}{gen}{2} & \BEa{qwenboth}{gen}{4} \\
\bottomrule
\end{tabular}
\end{table}

\paragraph{Per-task counts}
\Cref{tab:supp-main,tab:supp-ablation} list the counts plotted in \Cref{fig:main,fig:analysis}.

\paragraph{Statistical uncertainty}
\Cref{tab:supp-main} also reports Wilson intervals for the mean success rates.
Under shift, the intervals of \PsiZero and of every exocentric method are disjoint, and \Method's interval lies mostly above that of \Sone.
Because all methods are evaluated on the same frozen scenes, paired tests on per-scene outcomes are more powerful than these unpaired intervals.

\begin{table}[!htb]
\centering
\caption{\textbf{T1 generalization by shift family} (successes out of 8 scenes per family) for the foresight variants of \Cref{fig:analysis} whose per-family results are available.}
\label{tab:supp-slices}
\small
\renewcommand{\arraystretch}{1.16}
\setlength{\tabcolsep}{7pt}
\begin{tabular}{ll cccccc}
\toprule
\sffamily\bfseries Teacher &
\sffamily\bfseries Views &
\sffamily\bfseries Layout &
\sffamily\bfseries Asset &
\sffamily\bfseries Material &
\sffamily\bfseries Light &
\sffamily\bfseries Comb. &
\sffamily\bfseries All \\
\midrule
DINOv3 & ego & \BEs{dinoego}{layout} & \BEs{dinoego}{asset} & \BEs{dinoego}{material} & \BEs{dinoego}{lighting} & \BEs{dinoego}{combined} & \BEs{dinoego}{total} \\
DINOv3 & exo & \BEs{dinoexo}{layout} & \BEs{dinoexo}{asset} & \BEs{dinoexo}{material} & \BEs{dinoexo}{lighting} & \BEs{dinoexo}{combined} & \BEs{dinoexo}{total} \\
Qwen3-VL & ego & \BEs{qwenego}{layout} & \BEs{qwenego}{asset} & \BEs{qwenego}{material} & \BEs{qwenego}{lighting} & \BEs{qwenego}{combined} & \BEs{qwenego}{total} \\
Qwen3-VL & exo & \BEs{qwenexo}{layout} & \BEs{qwenexo}{asset} & \BEs{qwenexo}{material} & \BEs{qwenexo}{lighting} & \BEs{qwenexo}{combined} & \BEs{qwenexo}{total} \\
Qwen3-VL & both & \BEs{qwenboth}{layout} & \BEs{qwenboth}{asset} & \BEs{qwenboth}{material} & \BEs{qwenboth}{lighting} & \BEs{qwenboth}{combined} & \BEs{qwenboth}{total} \\
\bottomrule
\end{tabular}
\end{table}

\begin{figure}[!htb]
\centering
\includegraphics[width=\textwidth]{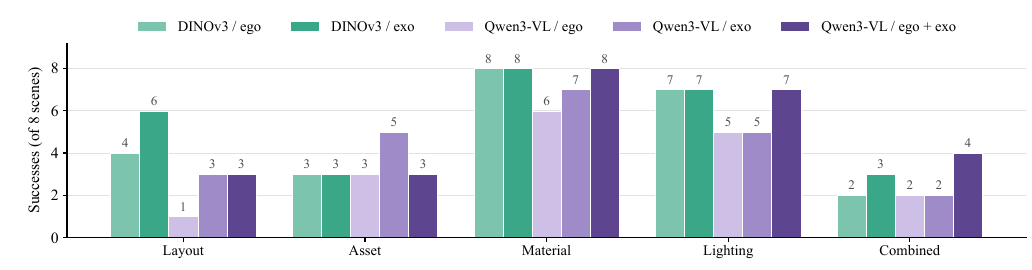}
\caption{\textbf{T1 generalization by shift family} for the foresight variants of \Cref{fig:analysis} whose per-family results are available.}
\label{fig:supp-slices}
\end{figure}

\paragraph{Shift families}
\Cref{tab:supp-slices} and \Cref{fig:supp-slices} break the T1 generalization results down by shift family.
Material and lighting shifts are the easiest; layout, asset and combined shifts are the hardest.
Exocentric DINOv3 foresight is the best single-view variant on layout shifts, which change the body--scene geometry along the route.

\begin{table}[!htb]
\centering
\caption{\textbf{Offline action errors over all five tasks} (final training step, sampled batches from the last 10 episodes of each task, mean over T1--T5; de-normalized MAE $\times10^{-3}$).}
\label{tab:supp-offline5}
\small
\renewcommand{\arraystretch}{1.16}
\setlength{\tabcolsep}{8pt}
\begin{tabular}{l ccccc}
\toprule
\sffamily\bfseries Method &
\sffamily\bfseries Hand &
\sffamily\bfseries Arm &
\sffamily\bfseries $v_x$ &
\sffamily\bfseries $v_y$ &
\sffamily\bfseries $v_{\text{yaw}}$ \\
\midrule
\PsiZero & \BEf{psi}{l1hand} & \BEf{psi}{l1arm} & \BEf{psi}{l1vx} & \BEf{psi}{l1vy} & \BEf{psi}{l1vyaw} \\
\Sone & \BEf{s1}{l1hand} & \BEf{s1}{l1arm} & \BEf{s1}{l1vx} & \BEf{s1}{l1vy} & \BEf{s1}{l1vyaw} \\
\midrule
\ywrowhighlight
\textbf{\Method} & \BEf{full}{l1hand} & \BEf{full}{l1arm} & \BEf{full}{l1vx} & \BEf{full}{l1vy} & \BEf{full}{l1vyaw} \\
\bottomrule
\end{tabular}
\end{table}

\paragraph{Offline errors across tasks}
\Cref{tab:supp-offline5} compares the egocentric baseline, \Sone and \Method offline.
The exocentric view mainly reduces the yaw-rate error (by about $2.8\times$), consistent with heading decisions that depend on the body--scene relation; hand errors are nearly identical, and the egocentric baseline has a slightly lower arm error.
The foresight expert of \Method reaches an offline cosine similarity of \BEf{cosego}{mean} (ego) and \BEf{cosexo}{mean} (exo) to the 0.6\,s-ahead DINOv3 features at the final training step, averaged over the five tasks (per task: ego \BEf{cosego}{1}, \BEf{cosego}{2}, \BEf{cosego}{3}, \BEf{cosego}{4}, \BEf{cosego}{5}; exo \BEf{cosexo}{1}, \BEf{cosexo}{2}, \BEf{cosexo}{3}, \BEf{cosexo}{4}, \BEf{cosexo}{5}).

\section{The foresight expert is removable}
\label{sec:supp-prune}

We verify the deployment claim of \Cref{sec:foresight} with three tests, run for each foresight configuration (DINOv3 targets on both views or on the egocentric view only, and Qwen3-VL targets on both views).
(i)~\emph{Target tampering}: shuffling the future frames across the batch and rescaling them, or omitting them altogether, leaves the generated action chunks bit-identical under fixed noise.
(ii)~\emph{Stripping}: deleting $F_\omega$ from the model leaves the generated action chunks bit-identical.
(iii)~\emph{Loading}: after the weights of $F_\omega$ are removed (95 tensors with two supervised views, 52 with one), the checkpoint loads into the deployment configuration, which does not build $F_\omega$, with zero missing and zero unexpected parameters.
Together these show that no part of the action path reads $F_\omega$ or the future frames, so foresight acts on the policy only through the trained weights of the adapter and the exocentric projection.

\section{Algorithms}
\label{sec:supp-algo}

\begin{algorithm}[!htb]
\caption{\Method training step}
\label{alg:train}
\begin{algorithmic}[1]
\Require batch of $(I^{\ego}_t,I^{\exo}_t,\ell,\mathbf s_t,\mathbf a_t)$ with future frames $I^{v}_{t+\Delta t}$ and validity masks
\State $Z^{\ego}_t\gets E_{\text{VLM}}(I^{\ego}_t,\ell)$; \ $Z^{\exo}_t\gets P_\psi(\Dino(I^{\exo}_t))$ \Comment{encoders frozen}
\State $\widetilde Z^{\ego}_t\gets Z^{\ego}_t+W_\alpha\LN(Z^{\ego}_t)$ \Comment{exo masked w.p.\ 0.1}
\State sample $\sigma$, $\boldsymbol\epsilon$, RTC prefix; \ $\mathbf a^\sigma_t\gets(1-\sigma)\mathbf a_t+\sigma\boldsymbol\epsilon$, with $\mathbf a_t$ normalized
\State $\Lact\gets\|v_\theta([\widetilde Z^{\ego}_t;Z^{\exo}_t],\mathbf s_t,\mathbf a^\sigma_t,\sigma)-(\boldsymbol\epsilon-\mathbf a_t)\|^2$ \Comment{\Cref{eq:act}}
\State $\widehat Z^{v}_{t+\Delta t}\gets\sg[\Dino(I^{v}_{t+\Delta t})]$ for $v\in\{\ego,\exo\}$
\State $\Lfut\gets$ masked mean of $1-\cos\big(\widehat Z^{v}_{t+\Delta t},F^{v}_\omega([\widetilde Z^{\ego}_t;Z^{\exo}_t],\mathbf s_t)\big)$ \Comment{\Cref{eq:fut}}
\State back-propagate $\Lact+\lambda(s)\Lfut$ \Comment{$\nabla_\theta\Lfut=\mathbf 0$}
\end{algorithmic}
\end{algorithm}

\begin{algorithm}[!htb]
\caption{\Method deployment}
\label{alg:deploy}
\begin{algorithmic}[1]
\State delete $F_\omega$ \Comment{no future frames are needed}
\Loop
\State observe $I^{\ego}_t$, $I^{\exo}_t$, $\mathbf s_t$; \ compute $\widetilde Z^{\ego}_t$ and $Z^{\exo}_t$ as in training
\State integrate $v_\theta$ in 10 Euler steps from $\sigma=1$ to $0$, with the RTC prefix of the previous chunk, and denormalize to obtain~$\mathbf a_t$
\State execute the first 15 commands through AMO
\EndLoop
\end{algorithmic}
\end{algorithm}

\Cref{alg:train,alg:deploy} summarize training and deployment.
Because $F_\omega$ reads only the observation interface, a single backward pass of the total loss suffices: the gradient of $\Lfut$ with respect to the action expert is zero by construction.

\FloatBarrier
\section{Further discussion}
\label{sec:supp-discussion}

\vspace{0.15em}\noindent{\normalfont\normalsize\sffamily\bfseries\color{yinwangdark}Why not feed the predicted future to the policy?}\hspace{0.65em}%
An alternative is to inject the output of $F_\omega$ into the action expert as extra tokens.
This makes $F_\omega$ part of the inference graph, so it can no longer be removed, and it lets the action expert rely on its own forecasts, whose errors may compound in closed loop.
\Method instead uses foresight only to shape the representation the policy reads.

\paragraph{Limitations}
Beyond the limitations in \Cref{sec:conclusion}, the exocentric camera is calibrated and fixed in all experiments, the deployed model retains the 4.2M-parameter adapter trained with the foresight loss, and the benchmark covers one robot and five tasks in simulation.

\paragraph{Broader impact}
More reliable humanoid loco-manipulation can assist logistics and household work.
Exocentric cameras observe the environment and possibly people in it; deployments should follow privacy regulations and inform the people being recorded.

\end{document}